\documentclass[twoside,11pt]{article}
\usepackage{jmlr2e}

\usepackage{caption}
\usepackage{subcaption}
\usepackage{graphicx}              
\usepackage{amsmath}               
\usepackage{amsfonts}    
\usepackage{siunitx}
\newcommand{\bd}[1]{\mathbf{#1}}  
\newcommand{\RR}{\mathbb{R}}      
\newcommand{\ZZ}{\mathbb{Z}}      

\usepackage[ruled,vlined, linesnumbered]{algorithm2e}

\usepackage{lastpage}

\begin{document}


\title{The Decoupled Extended Kalman Filter for Dynamic Exponential-Family Factorization Models}


\author{\name Carlos A. G\'{o}mez-Uribe \\ 
\email cgomez@alum.mit.edu \\
\AND
\name Brian Karrer \\
\email briankarrer@fb.com \\
\addr Facebook, Menlo Park, California, USA
}

\editor{}


\maketitle

\begin{abstract}%
Motivated by the needs of online large-scale recommender systems, we specialize the decoupled extended Kalman filter (DEKF) to factorization models, including factorization machines, matrix and tensor factorization, and illustrate the effectiveness of the approach through numerical experiments on synthetic and on real-world data.  Online learning of model parameters through the DEKF makes factorization models more broadly useful by (i) allowing for more flexible observations through the entire exponential family, (ii) modeling parameter drift, and (iii) producing parameter uncertainty estimates that can enable explore/exploit and other applications.  We use a different parameter dynamics than the standard DEKF, allowing parameter drift while encouraging reasonable values. We also present an alternate derivation of the extended Kalman filter and DEKF that highlights the role of the Fisher information matrix in the EKF.
\end{abstract}

\begin{keywords}
  approximate online inference, Kalman filter, matrix factorization, factorization machines, explore exploit.
\end{keywords}

\section{Introduction}
Today there are many examples of large-scale recommender systems that serve hundreds of millions of people every day, e.g., Netflix, Spotify, or Amazon. The scale is such that computationally efficient approaches are necessary. In addition, product preferences can change over time so ideal methods should naturally learn and respond to such changes. Lastly, an individual may only try a service a few times before deciding whether to continue to interact with it, e.g., see \citet{gomez2015netflix}. Learning an individual's preferences quickly is then also important. One useful strategy to address this coldstart situation involves explore-exploit strategies that rely on uncertainty estimates. The obvious need for computationally efficient methods to generate recommendations in situations with time-varying preferences and that can enable explore-exploit strategies is the main motivation for this work. The method we describe here extends matrix factorization, perhaps the most popular approach to recommendations, and other factorization models to meet these requirements: it makes these approaches online and dynamic, and provides uncertainty estimates to enable explore-exploit strategies. In addition, we develop our method to naturally handle a wide range of data types (e.g., positive integers, binary outcomes, or continuous outcomes) by working directly with the exponential family.

Our method utilizes Kalman filtering.  The Kalman filter (KF) was initially introduced in \citet{kalman1960new} for state estimation in linear systems driven by Gaussian noise, and with observations that depend linearly on the state and on additional Gaussian noise. The KF iteratively computes the exact posterior of the state as new observations become available. Many variants of the KF have since been developed and applied to a wide variety of models, e.g., for parameter learning. See \citet{simon2006optimal} and \citet{haykin2001kalman} for good overviews of Kalman filters; the latter is focused on neural network applications.

Regression, matrix and tensor factorization, factorization machines, and many other statistical models can be viewed as variations of a general model with exponential family observations.  An approximate Gaussian posterior of the parameters for this general model can be learned online, even when the parameters drift over time, through a KF called the extended Kalman filter (EKF), developed to handle non-Gaussian observations.  However, maintaining a full covariance matrix of the parameters, as prescribed by the EKF, can often be prohibitive in terms of memory and computation.  The decoupled EKF (DEKF) can alleviate this limitation.

The DEKF was introduced in \citet{puskorius1991decoupled} to train neural networks. It approximates the covariance matrix of the parameters in the EKF as block-diagonal. We will argue that this approximation is particularly relevant for models with a large number of parameters, such as factorization models, where only a relatively small subset of them is relevant to any given observation. Developing and applying the DEKF to factorization models has not been done before, and is the main contribution of this paper.  Specifically, we assume that model parameters can be naturally grouped into subsets we call \textit{entities}\footnote{These subsets are called \textit{nodes} in the original DEKF paper, but we find entity more descriptive for factorization models.},
such that few entities are involved in each observation.  E.g., in matrix factorization  exactly two subsets of parameters define each observation, those for the user and the item interacting, so we can let each user and each item correspond to an entity.

When a new observation arrives, we show the DEKF only requires updating the parameters of entities involved in the new observation. This leads to a particularly efficient implementation of the DEKF for factorization models.  Because the DEKF produces a posterior distribution of the parameters, it also enables applications that require uncertainty estimates, e.g., where explore/exploit trade-offs are important. For example, we show that the DEKF enables Thompson sampling in factorization models.

The DEKF we present here is different from the standard DEKF in several ways. First, we specialize it to exponential family models, motivated by models with typically few entities per observation. Second, the standard DEKF was formulated for static parameters, or for parameters that undergo a simple random walk. The latter choice can result in parameter values that become too large and lead to badly behaved models. Here, we consider parameter dynamics that allow for parameter drift while encouraging reasonable values.  Modeling parameter drift can be desirable in situations where the underlying data is non-stationary, as is often the case in recommender systems, where user preferences and item popularities can change over time. To keep our paper self-contained, we assume no familiarity with Kalman filtering.

The rest of this paper is organized as follows. Section \ref{sec:modelSetup} introduces the general model we study, and describes several kinds of factorization models as special cases. Section \ref{sec:mainAlgo} derives and describes our DEKF for factorization models with exponential family observations.  We then discuss connections of the EKF and DEKF to other related methods.  Section \ref{sec:numericalResults} describes numerical results on simulated and on real data, obtained from the application of our DEKF to a variety of models for the tasks of prediction and of reward maximization (explore/exploit). Section \ref{sec:discussion} concludes with a discussion about limitations, and suggests possible research directions.

\section{Dynamic Exponential Family Factorization Models}
\label{sec:modelSetup}
We consider a model with parameters at time $t$ denoted by $\theta_t \in \RR^k$. The parameters can be grouped into $n$ disjoint subsets $\xi_{i,t} \in \RR^{k_i}$, one per entity such as a user or an item, i.e., so that $\sum_{i=1}^n k_i = k$. We allow the parameters to drift over time, and model the observations $y_t \in \RR^d$ through the exponential family.  Matrices, with the exception of those defined implicitly through differentiation, are denoted by upper-case and bolded symbols. The generative model is:

\begin{enumerate}
    \item Initialize parameters.
    \begin{enumerate}
        \item Initialize \textit{reference vectors}: 
    \begin{align}
     r_i \sim 
    \mathcal{N}\big(\pi_{i}, \bd{\Pi}_i \big),  \label{eq:initRef} 
    \end{align}
    for $i = 1, \ldots, n,$ and with mean and covariance $\pi_{i}$ and $\bd{\Pi}_i$ assumed known.
        \item Initialize \textit{entity vectors}:
    \begin{align}
     \xi_{i,0} \sim \mathcal{N}\big(r_{i},  (1-\alpha_i^2)^{-1} \bd{\Omega}_i \big), \label{eq:initPars}    
    \end{align}
     for $i = 1, \ldots, n.$ The constants $\alpha_i$ and matrices $\bd{\Omega}_i$ are defined below. 
    \end{enumerate}
    \item Generate observations. For each time step $t > 0:$
        \begin{enumerate}
            \item Evolve parameters:
            \begin{align}
\xi_{i,t}  =& \alpha_i (\xi_{i,t-1} - r_{i}) + r_{i} + \omega_{i,t}.
\label{eq:dynamics} \end{align}
Here, $0 < \alpha_i \leq 1$ is a known constant called the memory parameter, and $\omega_{i,t} \sim \mathcal{N}\big(0, \bd{\Omega}_i\big)$ is the driving Gaussian noise, with known covariance $\bd{\Omega}_i.$
            \item Sample observation $y_t \in \RR^d$ from the natural exponential family\footnote{The exponential family is often defined using $T(y)$ instead of $y$ where $T(y)$ indicates the vector of sufficient statistics for an underlying vector of observations $y$.  To avoid additional notation, we consider our observation vector $y$ to just be the vector of sufficient statistics.}, with log-likelihood
            \begin{align} 
l(y_t) = \log{P(y_t)} = \eta' \bd{\Phi}^{-1} y_t - b(\eta, \bd{\Phi}) + c(y_t,\bd{\Phi}). \label{eq:logL}
\end{align} Here, $\eta \in \RR^d$ is the natural parameter of the distribution, a known function of $\xi_{i,t}$, and in some cases, of context $x_t.$ E.g., $x_t$ can be the predictors in a regression model or factorization machine, or the indices corresponding to the user and item involved in an observation for matrix factorization. The natural parameter $\eta$ is the connection between the model parameters we seek to estimate and the observations. Also, $\bd{\Phi} \in \RR^{d \times d}$ is a symmetric positive definite matrix that is a known nuisance parameter. The functions $b()$ and $c()$ depend on the specific member of the exponential family chosen for the model. Importantly, in a typical factorization model, very few entities are involved in each observation, i.e., $\eta$ is an explicit function of very few entities, e.g., one user and one item in matrix factorization. The symbol $'$ denotes the vector or matrix transpose operation.
        \end{enumerate}
\end{enumerate}
 As we will see, the model above includes regression and factorization models with static or dynamic parameters. In factorization models, typically $d \ll k.$ 
The observation model in Equation \ref{eq:logL} is a generalization of the Generalized Linear Model \citep[or GLM, see][]{hastie2017generalized, nelder1972generalized} based on a moderately more complex mapping between the model parameters and the parameters of the distribution that generates the observations in order to handle factorization models.\footnote{If the mapping from parameters to signal was arbitrary, this would be the Generalized Non-linear Model, but factorization models only require multi-linear maps.} Typically the nuisance parameter is the identity matrix, though in linear regression with known covariance, $\bd{\Phi}$ is the covariance of the observations.

Our goal is to estimate the distribution of $\theta_t,$ or equivalently, of $\xi_{i,t}$ for all entities $i$, given all the observations up to time $t$ in an online fashion. This estimation problem is generally analytically intractable, but we obtain approximate algorithms through Kalman filtering. A Kalman filter (KF) has two distinct steps that need to be performed in each timestep: a prediction of the parameters in light of their dynamics, and an update step that incorporates the information from the latest observation. To simplify the exposition, we first consider the special case where the model parameters are static, i.e., where $\bd{\Omega}_i = \bd{0}$ and $\alpha_i = 1$ for all $i$, so Equation \ref{eq:dynamics} simply becomes $\xi_{i,t} = \xi_{i,t-1}.$ The reference vectors then only serve to initialize the model parameters through Equation \ref{eq:initPars}, and the KF only consists of an update step that depends strongly on the observation model. We consider the more general model with dynamic parameters again in Section \ref{sec:dynamics}.

We denote the mean and covariance of $y_t$ given $\eta$ by $\mu_y(\eta)$ and $\bd{\Sigma}_y(\eta)$
, though we may omit the dependence on $\eta$ for improved readability. We also often omit the time subscript of $y,$ $\theta,$ $x,$ and other time-varying quantities for similar reasons. For distributions in the form of Equation \ref{eq:logL}, it can be shown that 
\begin{align} 
\mu_y(\eta) = & \bd{\Phi} \frac{\partial b}{\partial \eta}' = h(\eta), \label{eq:meanExp} \\
\bd{\Sigma}_y(\eta) = & \bd{\Phi} \frac{\partial^2 b}{\partial \eta^2} \bd{\Phi} = \frac{\partial h}{\partial \eta} \bd{\Phi}, \label{eq:varExp}
\end{align}
where $h(\eta)$, defined in the first equation, is called the \textit{response function}. Throughout our paper, our notation for vector and matrix derivatives is consistent with the notation of tensor calculus, which often results in the transposed vectors and matrices of other notations. E.g., here $\frac{\partial b}{\partial \eta}$ is a row vector in $\RR^d$, and  $\frac{\partial \eta}{\partial \theta}$ is a $d$-by-$k$ matrix. To connect the observations to the model parameters, we assume that $\eta$ is a deterministic and possibly non-linear function of $\theta$, with finite second derivatives. Often, $\eta$ is also a function of context denoted by $x$.  It is typical and helpful to think of an intermediate and simple function $\lambda$ of $\theta$ and $x$ that the natural parameter is a function of, i.e.,  $\eta = \eta(\lambda(\theta, x))$. This intermediate function $\lambda$ is called \textit{the signal}, and outputs values in $\RR^d$.  To avoid notation clutter, we suppress all dependencies on $x$. We will need to evaluate the mean and covariance of $y$ for specific values of $\theta \in \RR^k$. Abusing notation for improved readability, we will write $h(\theta)$ and $\bd{\Sigma}_y(\theta )$ instead of $h(\eta(\theta))$ and $\bd{\Sigma}_y(\eta(\theta))$ to denote the mean and covariance of $y$ at a specific value of $\theta$. 

The model also needs an invertible function called the \textit{link function}\footnote{Our definition of link function is often referred to as the inverse link function since it maps the signal to mean as opposed to the more traditional mapping of mean to signal.} $g(\lambda)$ that maps the signal to $\mu_y=h(\eta),$ so $\eta=h^{-1}(g(\lambda))$.  Depending on the family, $\mu_y$ can have a restricted range of values (e.g. $\mu_y > 0$), and for ease of exposition, we only consider link functions that obey these ranges without restricting the signal.  A particularly useful choice for the link function is the \textit{canonical link function} ($g = h$) that makes $\lambda=\eta,$ and simplifies relevant mathematics. Because the specific distribution within the exponential family determines $h(\eta)$, different distributions have different canonical links.  We will write $g(\theta)$ rather than $g(\lambda(\theta))$ for improved readability. To summarize, $\theta$ determines $\eta,$ but only through the signal $\lambda$. Then $\eta$ determines the mean and covariance of $y$ via Equations \ref{eq:meanExp} and \ref{eq:varExp}. Table \ref{table:notation} summarizes for convenience the main notation introduced so far, as well as some symbols that are introduced later.

\begin{table}[ht]
\centering
\begin{tabular}{|c| c|} 
 \hline
 Symbol & Variable  \\ [0.5ex] 
 \hline
 $y$ & observation  \\ 
 $x$ & context  \\ 
 $\theta$ & model parameters \\
 $\lambda(\theta, x)$ & signal \\
 $\eta(\lambda)$ & natural parameter of $y$ \\
 $\bd{\Phi}$ & nuisance parameter of the observation \\
 $h(\theta)$ & response function; mean of $y$ given $\theta$ \\
 $\bd{\Sigma}_y(\theta)$ & covariance of $y$ given $\theta$ \\
 $\xi_i$ & parameters of entity $i$, with  $\theta'=[\xi'_1 \ \ldots \ \xi'_n] $ \\
 $e(\theta)$ & prediction error $y - h(\theta)$ \\
 $\bd{F(\theta)}$ & Fisher information matrix \\
 $\mu, \bd{\Sigma}$ & mean and covariance of $\theta$ \\
 $\mu_i, \bd{\Sigma}_i$ & mean and covariance of $\xi_i$ \\
 $l(y)$ & log-likelihood of $y$  \\
 $\omega_{i,t}$ & noise driving the dynamics of entity $i$ \\
 $\bd{\Omega}_i$ & covariance of $\omega_{i,t}$  \\
 $\alpha_i$ & memory of dynamics for entity $i$ \\
 $r_{i}$ & reference vector of entity $i$ \\
 $\pi_{i}, \bd{\Pi}_i$ & initial mean and covariance of $r_{i}$ \\
 $\rho_{i}, \bd{P}_{i}$ & current mean and covariance of $r_{i}$ \\
 $\bd{R}_{i}$ & current covariance between $r_{i}$ and $\xi_i$ \\[1ex] 
 \hline
\end{tabular}
\caption{Notation.}
\label{table:notation}
\end{table}
\subsection{Model Examples}
\label{sec:modelExamples}

Different important model classes only differ in the mapping from $\theta$ to $\lambda$ in the observation model. For example:

\begin{enumerate}
    
\item \textbf{The GLM.} It is obtained by setting $\lambda=\bd{X}'\theta$, where $\bd{X} \in \RR^{k \times d}$ is a matrix of predictors.  The EKF has already been applied to the GLM with dynamic parameters, e.g., in \citet{gomez2016online}. But the DEKF can further enable learning for GLM models with many parameters and sparse $\bd{X}$, e.g., for a matrix factorization model with known items.

\item \textbf{Matrix factorization (MF).}
Consider a set of entities referred to as users and items, each described by a vector in $\RR^a$ for some small $a \in \ZZ^+$, and let $\theta \in \RR^{an}$ consist of the stacking of all $n$ user and item vectors.  The signal in these models is quadratic in $\theta$, and is given by the dot-product of the user and item vectors involved in an observation. Sometimes user and item bias terms are added to the signal too. 

MF models typically assume that the observations are univariate Gaussian, or occasionally Bernoulli, e.g., see \citet{mnih2008probabilistic} and \citet{koren2009matrix}, so our setup generalizes these models to observations in other exponential family distributions that can be more natural for different kinds of data. In addition, applying the DEKF to these models allows for user and item vector drift, and enables explore/exploit applications.

\item \textbf{Tensor factorization (TF).}
The CANDECOMP / PARAFAC (CP) decomposition of an order-$q$ tensor, described in \citet{Kolda:2009}, has entities for each of the $q$ dimensions of the tensor. When $q=2$, the model is equivalent to MF with two kinds of entities, users and items. Each entity in a TF model is described by a vector in $\RR^a$ for a small $a \in \ZZ^+,$ and is associated with one of the $q$ modes, e.g., users for mode one and items for mode two when $q=2$. Similarly, $\theta$ consists of stacking all these vectors together. Each observation $y$ is univariate, and describes the interaction between $q$ entities, one per mode. Denote the corresponding entity vectors involved in the observation by $\xi_{1}, \ldots, \xi_{q}$. The signal is defined as $\lambda = \sum_{l=1}^{a} \big(\prod_{i=1}^{q} \xi_{i l}\big)$, where $\xi_{i l}$ is the $l$-th entry of $\xi_i$. Note that when $q=2$ the signal is the same as in MF models. Our setup offers similar advantages in TF models as in MF models: flexible observations, parameter drift, and uncertainty estimates.

\item \textbf{Factorization machines (FM).}
These models, introduced in \citet{rendle2010factorization}, typically have univariate responses, and include univariate regression, MF, and tensor models as special cases. 

Assume there are $n$ entities, e.g., user or items that can be involved in any of the observations, and let $x_i$ be non-zero only when entity $i$ is involved in the observation, with $x=[x_1 \ldots x_n]'$. Let $\xi_i$ be the parameters corresponding to entity $i.$  In a factorization machine (FM) of order $2$, $\xi'_i = [w_i \ v'_i],$ where $w_i \in \RR$ and $v_i \in \RR^{a_2},$ with $a_2$ a positive integer, so $\xi_i \in \RR^{a_2+1}$. Then the signal becomes
\begin{align}
    \lambda = w_o + \sum_{i=1}^n w_ix_i + 
    \sum_{i=1}^n\sum_{j=i+1}^n v'_i v_j x_ix_j. \label{eq:SignalFMorder2}
\end{align}
When $x$ has exactly two non-zero entries set to 1, then Equation \ref{eq:SignalFMorder2} becomes identical to the signal in MF, with a user, item and a general bias term.  Higher-order factorization machines are described in \citet{rendle2010factorization}. FMs are learned via stochastic gradient descent, Markov Chain Monte Carlo, or alternating least squares or coordinate ascent \citep{rendle2012factorization}. Our treatment extends FMs beyond Bernoulli and Gaussian observations, allows for dynamic parameters, and provides parameter uncertainty estimates.
\end{enumerate}

\section{The Decoupled Extended Kalman Filter}
\label{sec:mainAlgo}
The EKF is a variant of the KF for non-linear dynamics and non-linear observations that results in approximate estimates of the state.  Like the standard KF, the EKF consists of an update step that incorporates a new observation into the parameter estimates, and a predict step that accounts for parameter dynamics. We describe the update step for the EKF next, and then show how this step simplifies in the DEKF. 

\subsection{The EKF Update Step}

We assume that at time $t$ but before $y_t$ is observed, $\theta \sim \mathcal{N}(\mu, \bd{\Sigma}),$ i.e., that the parameters have a Gaussian prior. 
The EKF computes an approximate Gaussian posterior for the parameters $\theta | y \sim \mathcal{N}(\mu_{\text{new}}, \bd{\Sigma}_{\text{new}}),$ where we omit the time subscript of $y.$  First, define the auxiliary matrix function
 \begin{align} 
\bd{B}(\theta) = &\bd{\Phi}^{-1} \bd{\Sigma}_y(\theta)    \bd{\Phi}^{-1} \frac{\partial \eta}{\partial \theta} \bd{\Sigma}   \frac{\partial \eta'}{\partial \theta}. \nonumber 
\end{align}
Given a value of $\theta$, $\bd{B}(\theta) \in \RR^{d \times d}$.  The mean and covariance of the approximate Gaussian posterior are then found via:
\begin{align} 
\mu_{\text{new}} = &\mu + \bd{\Sigma} \frac{\partial \eta'}{\partial \theta} |_{\mu}
\biggr[\bd{I} + \bd{B}(\mu) \biggr]^{-1} 
\bd{\Phi}^{-1} \biggr( y - h(\mu)\biggr),
\label{eq:meanUpdate} \\
\bd{\Sigma}_{\text{new}} = & \bd{\Sigma} - \bd{\Sigma} \frac{\partial \eta'}{\partial \theta} |_{\mu} 
\biggr[\bd{I} + \bd{B}(\mu) \biggr]^{-1} 
\bd{\Phi}^{-1}\bd{\Sigma}_y(\mu) \bd{\Phi}^{-1} \frac{\partial \eta}{\partial \theta} |_{\mu}   \bd{\Sigma}.
\label{eq:varUpdate}
\end{align}
Here $\frac{\partial \eta}{\partial \theta} |_{\mu}$ denotes $\frac{\partial \eta}{\partial \theta}$ evaluated at $\theta=\mu$, and we use that notation elsewhere for some function evaluations. Note that the matrix in the square brackets above, whose inverse is needed, is only of size $d$-by-$d$. Also, we see that the update to the mean in Equation \ref{eq:meanUpdate} is proportional to the error $e(\mu)=y - h(\mu)$.  Applying these equations to a specific model requires specifying the distribution of the observation, and the link function, to determine $\bd{\Phi}$, $\bd{\Sigma}_y(\mu)$, $h(\mu)$, and $\frac{\partial \eta}{\partial \lambda}.$ The latter is needed to compute $\frac{\partial \eta}{\partial \theta} = \frac{\partial \eta}{\partial \lambda} \frac{\partial \lambda}{\partial \theta}$.  The last quantity, $\frac{\partial \lambda}{\partial \theta},$ comes from the specific model being used, e.g., regression, MF, etc. 

A reader familiar with the extended Kalman filter may find it difficult to map the above expressions onto the standard EKF expressions.  To clarify the relationship, we convert from the exponential family's canonical to mean parameterization.  To do so, recall that $\mu_y = g(\lambda) = h(\eta)$.  Thus $\frac{\partial \eta}{\partial \theta}= (\frac{\partial h}{\partial \eta})^{-1} \frac{\partial g}{\partial \theta}$, and applying Eq.~\ref{eq:varExp}, $\frac{\partial \eta}{\partial \theta} = \bd{\Phi} \bd{\Sigma}_y^{-1} \frac{\partial g}{\partial \theta}$.  Inserting this relationship and simplifying gives the familiar update equations:
\begin{align} 
\mu_{\text{new}} = &\mu + \bd{\Sigma} \frac{\partial g'}{\partial \theta} |_{\mu}
\biggr[\bd{\Sigma}_y(\mu) + \frac{\partial g}{\partial \theta}|_{\mu} \bd{\Sigma}   \frac{\partial g'}{\partial \theta}|_{\mu} \biggr]^{-1} 
\biggr( y - g(\mu)\biggr),
\label{eq:meanUpdateMean} \\
\bd{\Sigma}_{\text{new}} = & \bd{\Sigma} - \bd{\Sigma} \frac{\partial g'}{\partial \theta} |_{\mu}
\biggr[\bd{\Sigma}_y(\mu) + \frac{\partial g}{\partial \theta}|_{\mu} \bd{\Sigma}   \frac{\partial g'}{\partial \theta}|_{\mu} \biggr]^{-1} \frac{\partial g}{\partial \theta} |_{\mu} \bd{\Sigma}.
\label{eq:varUpdateMean}
\end{align}

\subsubsection{Derivation}
A standard derivation of the EKF proceeds as follows: first, $y$ is approximated as a Gaussian according to 
$y \sim \mathcal{N}\big(h(\theta), \bd{\Sigma}_y(\mu)\big).$ Notice that the variance is evaluated at the mean of the prior, while the mean is allowed to depend on $\theta.$  To make the log-likelihood $l(y)$ a quadratic function of $\theta,$ $h(\theta)$ is approximated through a first-order Taylor expansion around $\mu$. We present an alternative derivation of the EKF update step for our general model that brings connections to other methods and statistical concepts more directly. This derivation directly illustrates why the DEKF is particularly appropriate for factorization models. 

We start by approximating $ l(y )$ as a quadratic function of $\theta$ through a second-order Taylor expansion about the prior mean $\mu$. We then take the expectation of the corresponding Hessian over the distribution of $y$ given $\eta$ to guarantee that the covariance matrix remains positive definite. Lastly, we do some algebra to obtain the desired EKF equations. In the special case of Gaussian observations and linear response function, the EKF approximations become equalities, and the update step of the EKF is identical to that of the KF.

To start, we note that
\begin{align} 
\frac{\partial l(y)'}{\partial \theta} = & \frac{\partial \eta'}{\partial \theta} \frac{\partial l(y)'}{\partial \eta} = \frac{\partial \eta'}{\partial \theta} \bd{\Phi}^{-1} e(\theta),
\nonumber
\end{align}
where $\frac{\partial \eta}{\partial \theta} = \frac{\partial \eta}{\partial \lambda} \frac{\partial \lambda}{\partial \theta} \in \RR^{d \times k}$ is the derivative of the natural parameter with respect to $\theta$.  The (conditional) Fisher information matrix plays a prominent role in our derivation. It is given by
\begin{align} 
\bd{F}(\theta) = & E_{y | \theta}\biggr[\frac{\partial l(y)'}{\partial \theta}\frac{\partial l(y)}{\partial \theta}\biggr] \nonumber\\
= & \frac{\partial \eta'}{\partial \theta} \bd{\Phi}^{-1} \bd{\Sigma}_y(\theta)    \bd{\Phi}^{-1} \frac{\partial \eta}{\partial \theta}, \label{eq:FisherHessian}
\end{align}
where the first equality is a definition, and the last equality is specific to our model assumptions.  We use the notation $E_{y | \theta}$ to emphasize that this expectation is over samples of $y$ from the statistical model with parameters $\theta$.\footnote{Recall that the natural parameter $\eta$, through the signal $\lambda$, can be a function of the context $x$ that accompanied the observation $y$.  The true Fisher information matrix is hence an average over the unknown distribution of contexts $x$ and over the model distribution for $y$ given $x$ and $\theta$. The above Fisher information is the conditional Fisher information considered for a fixed context $x$.}

The Hessian of the log-likelihood is
\begin{align}
\frac{\partial^2 l(y)}{\partial \theta^2} = & \frac{\partial \eta'}{\partial \theta}   \frac{\partial }{\partial \theta} \biggr( \bd{\Phi}^{-1} e(\theta) \biggr) + \sum_{j=1}^d  \frac{\partial^2 \eta_j}{\partial \theta^2} \biggr[\bd{\Phi}^{-1} e(\theta)\biggr]_j \nonumber \\ 
= & - \bd{F}(\theta) + \sum_{j=1}^d  \frac{\partial^2 \eta_j}{\partial \theta^2} \biggr[\bd{\Phi}^{-1} e(\theta)\biggr]_j , \label{eq:hessian}
\end{align}
an explicit function of the Fisher information matrix. Here, $\big[\bd{\Phi}^{-1} e(\theta)\big]_j$ is just the $j$-th entry of the vector $\bd{\Phi}^{-1} e(\theta).$  The first term in the last equation is a negative definite matrix. The second term is not necessarily negative definite, and we will see below that this could result in invalid covariance matrices that are not positive definite.  To avoid this situation, in our second-order Taylor expansion, we will replace the Hessian $\frac{\partial^2 l(y)}{\partial \theta^2}$ in Equation \ref{eq:hessian} by its average over $y$ given $\eta$, i.e., by $-\bd{F}(\theta)$. This is consistent with Equation \ref{eq:hessian}, which uses $y$ only in the second term on the right, through $e(\theta)$, and the error averaged over $y$ given $\eta$ is zero.\footnote{For a regression model using the canonical link, this is not an approximation because the second term in Equation~\ref{eq:hessian} is zero.  In general, we have that $\frac{\partial \eta_l}{\partial \theta_i \theta_j}=\frac{\partial \lambda'}{\partial \theta_i}\frac{\partial^2 \eta_l}{\partial \lambda^2}\frac{\partial \lambda}{\partial \theta_j}  + \frac{\partial \eta_l}{\partial \lambda}\frac{\partial \lambda}{\partial \theta_i \partial \theta_j}.$  For any model where the canonical link is used, the signal \textit{is} the natural parameter because of the canonical link, so $\frac{\partial^2 \eta_l}{\partial \lambda^2}=\bd{0}.$  For regression models, the second term is also zero, because the signal is linear in $\theta$. So for regression models with the canonical link, the Hessian is identical to the negative Fisher information matrix.}

Combining these results we obtain our second-order approximation of the log-likelihood about the prior mean $\mu$: 
\begin{align}
l(y) \approx & l(y, \mu) + \frac{\partial l(y)}{\partial \theta} |_{\mu} \big(\theta - \mu \big) + \frac{1}{2}
\big(\theta - \mu \big)'   E_{y | \mu}\biggr[\frac{\partial^2 l(y)}{\partial \theta^2}  |_{\mu} \biggr]  \big(\theta - \mu \big) \nonumber \\
=&  l(y, \mu) + e(\mu)' \bd{\Phi}^{-1}   \frac{\partial \eta}{\partial \theta}  |_{\mu} \big(\theta - \mu \big) - \frac{1}{2}
\big(\theta - \mu \big)'\bd{F}(\mu)
\big(\theta - \mu\big). \nonumber
\end{align}
Plugging this approximation into
\begin{align} 
\log{P(\theta | y)} \propto & \ 
 \log{P(\theta)} +  l(y ), \label{eq:bayes}
\end{align}
as well as writing the Gaussian prior of $\theta$, while dropping terms independent of $\theta$ yields
\begin{align} 
\log{P(\theta | y)} \propto & -\frac{1}{2}
\big(\theta - \mu \big)'
\biggr( \bd{\Sigma}^{-1} + \bd{F}(\mu) \biggr)  
\big(\theta - \mu\big) +  e(\mu)' \bd{\Phi}^{-1}   \frac{\partial \eta}{\partial \theta}  |_{\mu} \big(\theta - \mu \big) \nonumber \\
 = & -\frac{1}{2}
\big(\theta - \mu - \bd{\delta} \big)'
\bd{\Sigma}^{-1}_{\text{new}}
\big(\theta - \mu -\bd{\delta} \big), \label{eq:updateDer}
\end{align}
with 
 \begin{align} 
\bd{\Sigma}^{-1}_{\text{new}} = &  \bd{\Sigma}^{-1} + \bd{F}(\mu) , \label{eq:varUpdateDer} \\
\bd{\delta} = & \bd{\Sigma}_{\text{new}} \frac{\partial \eta'}{\partial \theta} |_{\mu} \bd{\Phi}^{-1}  e(\mu). \label{eq:meanUpdateDer}
\end{align}
The last equality in Equation \ref{eq:updateDer} is obtained by completing squares. The result shows that the approximate posterior distribution is Gaussian with mean $\mu + \bd{\delta}$ and covariance $\bd{\Sigma}_{\text{new}}$.

The EKF covariance update then follows from applying the Woodbury identity \citep[see][sec. 3.2]{petersen2008matrix} to Equation \ref{eq:varUpdateDer}, and some re-arrangement. Plugging the updated covariance into Equation \ref{eq:meanUpdateDer} yields the EKF mean update, also after some re-arrangement.

\subsection{The DEKF Update Step}
The DEKF makes the additional assumption that the prior and all posterior covariances of the model parameters are block-diagonal with the same block structure, with each block corresponding to one entity. We show below that this assumption implies that only the entities in an observation need to be updated when that observation is processed, i.e., we only need to to apply Equations \ref{eq:meanUpdate} and \ref{eq:varUpdate} for the entities involved in the last observation.  The estimates for the rest of the parameters remain unchanged from their prior estimates.  Equally important, only means and covariances associated with the entities in the observation are needed to compute the updates.  A third important consequence of the DEKF assumption is that we can add new entities as they appear, which can be required for some online settings, e.g., in recommender systems where new users and items appear all the time.  The parameters for entities that have not been involved in any observations can just be appended into the set of parameters when the entity is first observed.

Consider the evaluation of Equations \ref{eq:meanUpdate} and \ref{eq:varUpdate}, both of which rely upon the computation of $\frac{\partial \eta}{\partial \theta} \bd{\Sigma}$ and $   \frac{\partial \eta}{\partial \theta} \bd{\Sigma}   \frac{\partial \eta'}{\partial \theta},$ evaluated at $\theta=\mu$. Without loss of generality, assume only the first $m$ entities are involved in the observation, so we have that
\begin{align}
    \frac{\partial \eta}{\partial \theta} = &
    \biggr[\frac{\partial \eta}{\partial \xi_1} \ldots \frac{\partial \eta}{\partial \xi_m} \ \bd{0} \biggr], \nonumber
\end{align}
where $\bd{0}$ is a matrix with entries set to zero of dimensions $d \times (k - \sum_{i=1}^m k_i)$. 
Combined with the block-diagonal structure of $\bd{\Sigma}$, this yields 
\begin{align}
 \frac{\partial \eta}{\partial \theta} \bd{\Sigma}  = & \biggr[ \frac{\partial \eta}{\partial \xi_1} \bd{\Sigma}_1
    \ldots \frac{\partial \eta}{\partial \xi_m} \bd{\Sigma}_m \ \bd{0} \biggr], \nonumber \\
    \frac{\partial \eta}{\partial \theta} \bd{\Sigma}   \frac{\partial \eta'}{\partial \theta} = &
    \sum_{i=1}^m \frac{\partial \eta}{\partial \xi_i}\bd{\Sigma}_i \frac{\partial \eta'}{\partial \xi_i}. \nonumber
\end{align}
where $\bd{0}$ is again defined to have the appropriate dimensions.  Substituting the first of these equations into the EKF update equations shows that only the $m$ entities involved in the observation are updated, whereas all others remain the same.  Examining the terms, we see that only means and covariances involved in the observation are used to compute the updates as well.

\begin{algorithm}[htb]
 \KwData{Observation $y$, context $x$, prior mean $\mu$, prior covariance $\bd{\Sigma}$}
 \KwResult{$\mu_{\text{new}}$, $\bd{\Sigma}_{\text{new}}$, updated in-place}
 Let $\xi_\lambda$ be the $m$ entities involved in the observation.
 Initialize $\bd{D} = \bd{0} \in \RR^{d \times d}$.
 
 \For{$i$ in $\xi_\lambda$} {
    $\bd{Q}_i = \bd{\Sigma}_i \frac{\partial \eta'}{\partial \xi_i}$
    
    $\bd{D} \mathrel{+}= \frac{\partial \eta}{\partial \xi_i} \bd{Q}_i $
 }
 Let $\bd{A} = \bd{\Phi}^{-1} \bd{\Sigma}_y(\mu) \bd{\Phi}^{-1}$, $\bd{B} = (\bd{I} + \bd{A}\bd{D})^{-1}$, $\bd{C} = \bd{B}\bd{A}$, and $f = \bd{B}\bd{\Phi}^{-1} \biggr( y - h(\mu)\biggr)$.
 
 \For{$i$ in $\xi_\lambda$} {
    $\mu_i \mathrel{+}= \bd{Q}_i f$
    
    $\bd{\Sigma}_{i} \mathrel{-}=  \bd{Q}_i \bd{C} \bd{Q}'_i$
 }
\Return{$\mu$, $\bd{\Sigma}$}
\caption{DEKF for models with static parameters.}
\label{algo:framework}
\end{algorithm}

Evaluation of the expressions above at $\theta = \mu$ leaves little extra work to compute the updated parameters $\mu_{\text{new}}$ and $\bd{\Sigma}_{\text{new}}.$ The resulting EKF posterior covariance $\bd{\Sigma}_{\text{new}},$ however, is typically not block-diagonal over the entities.  Letting $\bd{\Sigma}_{ij,\text{new}}$ denote the updated block for entities $i$ and $j$ in the observation, we have that
\begin{align}
    \bd{\Sigma}_{ij,\text{new}} = & \bd{\Sigma}_{ij} - \bd{\Sigma}_{i} \frac{\partial \eta'}{\partial \xi_i} |_{\mu} 
\biggr[\bd{I} + \bd{B}(\mu) \biggr]^{-1} 
\bd{\Phi}^{-1}\bd{\Sigma}_y(\mu) \bd{\Phi}^{-1} \frac{\partial \eta}{\partial \xi_j} |_{\mu} \bd{\Sigma}_j. \nonumber
\end{align}
The updated $\bd{\Sigma}_{ij,\text{new}}$ will generally be non-zero for any pair of entities $i$ and $j$ involved in the observation, even when $\bd{\Sigma}_{ij}=\bd{0}$.  To retain the desired block-diagonal covariance, the DEKF approximates the posterior by zeroing out any off-diagonal covariance blocks.  In practice, we simply never compute off-diagonal blocks.  This finishes the update step for the DEKF that reflects the new observation in the parameter estimates. For models with static parameters, the DEKF only has an update step, resulting in Algorithm~\ref{algo:framework}. The memory storage is $O(k^2)$ and the computation per observation is $O(k^2 + d^3)$ for the EKF. The DEKF, in contrast, is $O(\sum_{i=1}^{n} k_i^2)$ for storage and $O(\sum_{i \in \xi_\lambda} k_i^2 + d^3)$ for computation per observation, where $\xi_\lambda$ are the indices of the $m$ entities involved in the observation.  The reduction in both memory storage and computation for factorization models where the number of entities is large can thus be significant.

\subsection{The EKF covariance for factorization models}
\label{sec:ekf_covariance}

As shown, the EKF covariance update in Equation \ref{eq:varUpdate} will densify the covariance matrix, filling in covariance blocks across entities as observations accumulate. A good choice of entities for the DEKF will result in few, and small in dimension and in magnitude, $\bd{\Sigma}_{ij,\text{new}}$ off-diagonal blocks being non-zero in the full EKF update procedures.  We therefore suggest that reasonable entities to use, within the capabilities of available memory and computation, are commonly co-occurring non-zero parameter components of the gradient of the natural parameter.  Here we provide intuition on why that is a reasonable choice and note that it is trivial to identify such entities for factorization models.

We begin by rewriting the EKF inverse covariance update in Equation~\ref{eq:varUpdateDer} more explicitly.  We add an explicit observation index $t$ and display the dependence on each observation's context in the conditional Fisher information
\begin{align}
\bd{\Sigma}^{-1}_{t+1} = &  \bd{\Sigma}^{-1}_{t} + \bd{F}(\mu_t, x_t).
\end{align}
Iterating this for $T$ observations gives
\begin{align}
\bd{\Sigma}^{-1}_{T+1} = & \bd{\Sigma}^{-1}_0 + \sum_{t=1}^{T} \bd{F}(\mu_t, x_t),
\end{align}
where $\bd{\Sigma}^{-1}_0$ is the initial inverse prior covariance.  For large enough $T$, the first term can become irrelevant.  The second term has a non-zero contribution from observation $t$ only for matrix entries corresponding to parameters involved in the observation, i.e., entry $i,j$ of $\bd{F}(\mu_t, x_t)$ is not zero only if the gradients of the natural parameter $\eta$ with respect to parameters $i$ and $j$ are both non-zero for observation $t$. So roughly speaking the inverse covariance per observation for parameter $i$ and $j$ is proportional to the co-occurrence frequency of parameter $i$ and $j$ in observations.  Similarly, the inverse covariance per observation for parameter $i$ is roughly proportional to the marginal frequency of parameter $i$'s involvement in an observation.

We suggest to group subsets of parameters that have high co-occurrence in $\bd{F}(\mu_t, x_t)$ into entities.  The resulting entities will then correspond to blocks with substantially larger values in the inverse covariance per observation, because the co-occurrence frequency of parameters belonging to different entities is typically smaller than the within-entity frequency. This also indicates why a fully diagonal approximation to the covariance may be worse than the DEKF block-diagonal approximation: off-diagonal co-occurrence frequencies similar (or even equal) to the marginal frequencies would be ignored in its inverse.

\subsection{Parameter Dynamics}
\label{sec:dynamics}
We now consider the full model where parameters follow the dynamics of Equation \ref{eq:dynamics}. Parameter estimates need to be changed between observations to reflect these dynamics, resulting in the predict step of Kalman filtering. In typical Kalman filtering applications, the parameters (or state) are assumed to undergo known linear dynamics plus Gaussian noise according to
\begin{align}
\theta_{t} = \bd{G}_t \theta_{t-1} + u_t + \bd{\epsilon}_t. \nonumber
\end{align}
Here $\bd{\epsilon}_t$ is additive Gaussian noise, and the dynamics matrix $\bd{G}_t$ and the vector $u_t$ are known. In the EKF (and the original DEKF), the true dynamics are defined by non-linear functions, that are approximated through a first order Taylor expansion about the mean of the current posterior, resulting in essentially the same linear dynamics above. 

For our purposes, these dynamics are too general, since the parameters $\bd{G}_t$ and $u_t$ are typically unknown in machine learning applications.  We consider parameter dynamics here only as a means to incorporate data non-stationarity.  So we assume each entity $i$ evolves independently of the others according to Equation \ref{eq:dynamics}. There, the memory parameter $\alpha_i$ provides a form of regularization towards the reference vector $r_{i}.$

Our motivation for adding reference vectors and the memory parameter $\alpha_i$ is two-fold.  First, if $\alpha_i = 1$, the entity parameters undergo a random walk, and can accumulate a large covariance.  In MF models such a random walk often leads to user and item vectors that produce absurdly large signals. In contrast, Equation \ref{eq:dynamics} implies the steady-state distribution $\xi_{i} \sim \mathcal{N}\big(r_{i},  (1-\alpha_i^2)^{-1} \bd{\Omega}_i \big)$, which we use to initialize the entity vectors (Equation \ref{eq:initPars}). Second, these dynamics allow predicting a reasonable mean, i.e., the reference vector, for entities that have not been observed in a long time.  This is particularly relevant for factorization models, where entities may be observed infrequently.

\begin{algorithm}[!htb]
 \KwData{Observation $y$, context $x$, time $t$, prior mean $\mu$, prior covariance $\bd{\Sigma}$, last update time per entity $\tau$ }
 \KwResult{$\mu_{\text{new}}$, $\bd{\Sigma}_{\text{new}}$, $\tau_{\text{new}}$ updated in-place}
 Let $\xi_\lambda$ be the $m$ entities involved in the observation.
 
 \tcc{---Predict step---} 
 
 \For{$i$ in $\xi_\lambda$}{
    \uIf{entity $i$ exists}{
        $k = \tau_i$, $\tau_i = t$, $\mu_{i} = \alpha_i^{t-k} (\mu_{i} - \rho_{i}) + \rho_{i}$
        
        $\bd{\Sigma}_{i} = \frac{1-\alpha_i^{2(t-k)}}{1-\alpha_i^2} \bd{\Omega} + \alpha_i^{2(t-k)}\bd{\Sigma}_{i} + (\alpha_i^{2(t-k)} - 2\alpha_i^{t-k} + 1)\bd{P}_{i} + (\alpha_i^{t-k} - \alpha_i^{2(t-k)})(\bd{R}_{i} + \bd{R}_{i}')$
        
        $\bd{R}_{i} = \alpha_i^{t-k}\bd{R}_{i} + (1-\alpha_i^{t-k})\bd{P}_{i}$
    }
    \Else{
        $\tau_i = t$,
        $\mu_i = \pi_i$, $\rho_{i} = \pi_i$
        
        $\bd{\Sigma}_{i} = \bd{\Pi}_i + (1-\alpha_i^2)^{-1} \bd{\Omega}_i$, $\bd{R}_{i} = \bd{\Pi}_i$, $\bd{P}_{i} = \bd{\Pi}_i$
    }
 }
 
 \tcc{---Update step---}
 Initialize $\bd{D} = \bd{0} \in \RR^{d \times d}$
 
 \For{$i$ in $\xi_\lambda$} {
    $\bd{Q}_i = \bd{\Sigma}_i \frac{\partial \eta'}{\partial \xi_i}$, 
    $\bd{S}_i = \bd{R}_{i} \frac{\partial \eta'}{\partial \xi_i}$,
    $\bd{D} \mathrel{+}= \frac{\partial \eta}{\partial \xi_i} \bd{Q}_i $
 }
 
 Let $\bd{A} = \bd{\Phi}^{-1} \bd{\Sigma}_y(\mu) \bd{\Phi}^{-1}$, $\bd{B} = (\bd{I} + \bd{A}\bd{D})^{-1}$, $\bd{C} = \bd{B}\bd{A}$, and $f = \bd{B}\bd{\Phi}^{-1} \biggr( y - h(\mu)\biggr)$.
 
 \For{$i$ in $\xi_\lambda$} {
    $\mu_i \mathrel{+}= \bd{Q}_i f$,
    $\rho_{i} \mathrel{+}= \bd{S}_i f$
    
    $\bd{G}_i = \bd{C} \bd{Q}'_i$
    
    $\bd{\Sigma}_i \mathrel{-}=  \bd{Q}_i \bd{G}_i$,
    $\bd{R}_{i} \mathrel{-}=  \bd{S}_i \bd{G}_i$, 
    $\bd{P}_{i} \mathrel{-}= \bd{S}_i \bd{C} \bd{S}_{i}'$
 }
 
\Return{$\mu$, $\bd{\Sigma}$, $\tau$}
\caption{The DEKF optimized for dynamic factorization models.}
\label{algo:frameworkWithDrift}
\end{algorithm}

With parameter dynamics, $\theta$ includes both the reference and the entity vectors. We expand our notation to let $\rho_{i}$ denote the current mean of $r_i$,  $\bd{R}_{i}$ the covariance between $r_{i}$ and $\xi_i$, and $\bd{P}_{i}$ the covariance matrix of $r_i$.  An entity now refers to both the subset of current model parameters, $\xi_i$, and its associated reference vector $r_{i}$.  The DEKF posterior maintains a block-diagonal covariance over these augmented entities because $\bd{R}_{i}$ is generally non-zero for any entity $i$.

The update step in Algorithm~\ref{algo:framework} is still valid now that there are parameter dynamics, after replacing $\frac{\partial \eta}{\partial \xi_{i}}$ with a gradient with respect to the complete set of entity $i$'s parameters, both current and reference, i.e. $\left[\frac{\partial \eta}{\partial \xi_{i}}, \frac{\partial \eta}{\partial r_{i}}\right]$, and similarly replacing $\bd{\Sigma}_i$. However, this replacement is inefficient since the gradient of the log-likelihood with respect to the reference vectors is always zero, because $\frac{\partial \eta}{\partial r_{i}} = \bd{0}$. Our full variant DEKF, in  Algorithm~\ref{algo:frameworkWithDrift}, modifies the update step of Algorithm~\ref{algo:framework} to remove this inefficiency. In Algorithm~\ref{algo:frameworkWithDrift},  $\frac{\partial \eta}{\partial \xi_i}$ is just the gradient of $\eta$ with respect to the entity's current parameters $\xi_i$.

The main change in a Kalman filter when adding parameter dynamics is the presence of the predict step. Because the DEKF update step only requires means and covariances for the entities involved in the observation, we are only required to apply the predict step for those entities when observed.  In particular, the predict step can be applied immediately before the update step for the set of entities in an observation.  This is possible because our dynamics is completely independent across entities.  As opposed to laboriously maintaining a posterior over all parameters at time $t$, we can just maintain a \textit{lazy} posterior over each entity by recording only the most recent posterior for each entity, and the last time that entity was updated.  This is statistically identical to an inference procedure that would update the posterior for all entities at every time step.

Consider a particular entity $i$.  When we predict for this entity at time $t$, we first check whether we already have a past mean and covariance for the parameters of this entity.  If not, we assume the current parameters are drawn from the steady-state distribution of the dynamics and set the means and covariances to   
\begin{align}
\begin{bmatrix} \mu_i \\ \rho_{i} \end{bmatrix} = \begin{bmatrix} \pi_{i} \\ \pi_{i} \end{bmatrix}, \nonumber
\end{align}
and
\begin{align}
\begin{bmatrix} \bd{\Sigma}_i \\ \bd{R}_{i} \\ \bd{P}_{i} \end{bmatrix} = \begin{bmatrix} \bd{\Pi}_i + (1-\alpha_i^2)^{-1} \bd{\Omega}_i \\ \bd{\Pi}_i \\  \bd{\Pi}_i \end{bmatrix}. \nonumber
\end{align}
If entity $i$ has a posterior that was last updated at time $k$, we can write down the entire dynamics for the corresponding parameters between time $k$ and the current time $t$, as
 \begin{align}
\xi_{i,t}  =& \alpha_i^{t-k} (\xi_{i,k} - r_{i})+ r_{i} + \sum_{r=0}^{t-k-1}\alpha_i^{r} \omega_{i,r+k+1}. \nonumber
\end{align}
which implies we can directly update the entity's posterior at time $k$ to the posterior at time $t$. For the means, we have
\begin{align}
 \begin{bmatrix} \mu_{i,\text{new}} \\ \rho_{i,\text{new}} \end{bmatrix} = \begin{bmatrix}  \alpha_i^{t-k} (\mu_{i} - \rho_{i}) + \rho_{i} \\ \rho_{i} \end{bmatrix}. \nonumber
\end{align}
For the covariances, we have
\begin{align}
\begin{bmatrix} \bd{\Sigma}_{i,\text{new}} \\ \\ \bd{R}_{i,\text{new}} \\  \bd{P}_{i,\text{new}} \end{bmatrix} = \begin{bmatrix} \frac{1-\alpha_i^{2(t-k)}}{1-\alpha_i^2} \bd{\Omega} + \alpha_i^{2(t-k)}\bd{\Sigma}_{i} + (\alpha_i^{2(t-k)} - 2\alpha_i^{t-k} + 1)\bd{P}_{i} + \\ (\alpha_i^{t-k} - \alpha_i^{2(t-k)})(\bd{R}_{i} + \bd{R}_{i}') \\ \alpha_i^{t-k} \bd{R}_{i} + (1-\alpha_i^{t-k})\bd{P}_{i} \\ \bd{P}_{i} \end{bmatrix}. \nonumber
\end{align}
Because the predict step for entities can predict across any number of discrete time-steps with the same computational cost, our particular choice of entity dynamics allows us to incorporate parameter drift efficiently. We summarize the complete algorithm with the predict-update cycle in Algorithm~\ref{algo:frameworkWithDrift}.

\subsection{Model Examples}
As mentioned earlier, to apply the algorithm to a concrete model, we find $\frac{\partial \eta}{\partial \xi_i} = \frac{\partial \eta}{\partial \lambda} \frac{\partial \lambda}{\partial \xi_i},$ and substitute it into the procedures above. The first term, $\frac{\partial \eta}{\partial \lambda},$ follows from the choice of link function used. The second term, $\frac{\partial \lambda}{\partial \xi_i},$ is the gradient of the signal with respect to an entity, and depends on the model class. It is easy to write it down explicitly for many models, but can also be derived implicitly via automatic differentiation applied to the signal function. We consider the following models for simulations and analysis:

\begin{enumerate}
\item \textbf{Multivariate regression.}
The simplest model class we consider is regression, where $\lambda=\bd{X}'\theta$, so that $\frac{\partial \lambda}{\partial \theta} = \bd{X}' \in \RR^{d \times k}$, and $\frac{\partial \lambda}{\partial \xi_i} = \bd{X}_i',$ where $\bd{X}^{'}_i \in \RR^{d \times k_i}$ is the subset of $k_i$ columns of $\bd{X'}$ corresponding to entity $i$.
\footnote{A similar algorithm for the GLM was developed in \citet{gomez2016online}, but without entities and reference vectors, and working with the Hessian of the log-likelihood rather than with the Fisher information matrix (which coincide when the canonical link is used).}
\item \textbf{Matrix factorization.} Since $\lambda=\xi_u'\xi_v$, we have that
$\frac{\partial \lambda}{\partial \xi_u} = \xi_v$ and $
\frac{\partial \lambda}{\partial \xi_v} = \xi_u$. For all other entities, $\frac{\partial \lambda}{\partial \xi_i} = 0$.
\item \textbf{Tensor factorization.} Assume entities $1,\ldots, q$ are involved in the observation. We then have that
$\lambda = \sum_{l=1}^{a} (\prod_{i=1}^{q} \xi_{il}).$ The signal gradient is then 
$\frac{\partial \lambda}{\partial \xi_{i l}} = \prod\limits_{k=1, k \neq i }^{q} \xi_{kl}$ for $i=1,\ldots,q,$ and $\frac{\partial \lambda}{\partial \xi_i} = 0$ for other entities $i$.
\end{enumerate}

\subsection{Related Work}

\citet{ollivier2017online} shows that the EKF update step in Equations $\ref{eq:varUpdateDer}$ and $\ref{eq:meanUpdateDer}$ is equivalent to the computations in the online natural gradient algorithm \citep{Amari:1998} under the specific and commonly used choice of learning rate $1/(t+1),$ where $t$ is the number of observations.  However, even though the parameter updates in response to an observation are the same, the online natural gradient algorithm does not attempt to handle dynamic parameters. There have been more recent efforts, however, to approximate the matrix in the algorithm update through sparse graphical models \citep{grosse2015scaling}, and the use of Kronecker products \citep{martens2015optimizing}.

With exponential family observations, the Fisher information matrix is equivalent to the Generalized-Gauss-Newton (GGN) matrix in some circumstances \citep{Martens2014NewIA}.  This connects the online natural gradient to Hessian-Free optimization and Krylov Subspace Descent methods when applied to neural networks with exponential family observations \citep{Pascanu14revisitingnatural, Martens2014NewIA}. The TONGA algorithm was introduced in~\citet{NIPS2007_3234} for fitting neural networks, utilizing a block-diagonal low-rank approximation.  Claimed to perform online natural gradient, it was later shown~\citep{Pascanu14revisitingnatural} to be a related approach that used the outer-product of gradients evaluated at the observed $y$ (sometimes referred to as the empirical Fisher matrix), instead of the expectation over $y$ for the Fisher information as in Equation~\ref{eq:FisherHessian}. The empirical Fisher matrix is commonly used in adaptive stochastic gradient methods, including AdaGrad, RMSProp, and Adam among others, e.g., see~\citep{Martens2014NewIA}, where an argument is also made about why the Fisher information can be a better choice than its empirical counterpart.   Fitting neural networks using the Fisher information appeared earlier in~\citet{Kurita1994}, where an online block-diagonal approximation was considered as a Fisher scoring variant.

Another broad class of learning algorithms is Markov Chain Monte Carlo (MCMC), e.g., Gibbs sampling, Metropolis Hastings, etc. These algorithms generate sequences of parameter values rather than maintain a probabilistic model of the parameters. Recently, MCMC algorithms based on Langevin dynamics have been proposed that generate samples of the posterior distribution. The stochastic gradient Fisher scoring (SGFS) in \citet{ahn2012bayesian} is somewhat similar to our algorithm, and resembles online Fisher scoring driven by Gaussian noise. However, compared to our algorithm, it is not specifically online, does not maintain a distribution of the parameters, nor has been developed for entities.

\section{Numerical Results}
\label{sec:numericalResults}
We first apply the DEKF to simulated data, both with static and with dynamic parameters. For simplicity of exposition, we define a single observation model, and couple it to the model parameters through different signal definitions to obtain regression, matrix and tensor factorization models. Consider a stream of univariate binary observations $y_t$ and context $x_t$ provided at time $t$. We generate this stream according to the generative model described in Section~\ref{sec:modelSetup}. That is, we simulate the entity dynamics over each time step explicitly, and sample an observation $y_t$ by randomly selecting the entities involved, and possibly additional context. We learn the model parameters from this stream of observations via the EKF and DEKF algorithms. When Algorithm~\ref{algo:frameworkWithDrift}, that takes parameter dynamics into account, is applied for parameter learning, we use the true values for $\bd{\Omega}_i$ and $\alpha_i$ in it, i.e., we use the same values to generate the data and to learn the parameters. Similarly, the same values for the entity priors $\pi_i$ and $\bd{\Pi}_i$ are used for data generation and parameter learning.

We model the binary observations using the Bernoulli exponential family with the canonical link.  With this choice, as in logistic regression, the probability of an observation is $p_y(\eta) = \frac{e^{\eta}}{1+e^{\eta}}$.  So $h(\mu) = p_y(\mu),$ and the variance $\sigma^2_y$ is $p_y(\mu)(1-p_y(\mu))$.  With the canonical link, $\frac{\partial \eta}{\partial \lambda} = 1$.  Finally, the Bernoulli log-likelihood is $y \eta +\log\big(1-p_y\big)$, so $\bd{\Phi}=1$. The prior covariance $\bd{\Pi}_i$ for an arbitrary entity $i$ with $k_i$ entries was obtained as follows in every simulation. First, we construct a $k_i$ by $k_i$ matrix $\bd{U_1}$ by sampling its entries independently from the uniform distribution in $[0,1].$ Then, we obtain a positive definite matrix with non-negative entries by setting $\bd{U_2} = \bd{U_1}\bd{U_1'}/k_i^2$. We choose the latter because we want the entries of the initial parameter vectors to have positive correlation. We finally re-scale the matrix to have reasonable magnitude via  $\bd{\Pi}_i =\frac{s_p}{u} \bd{U_2}$, where $s_p$ is a typical value, to be specified later, that we want to achieve in the diagonal entities of the prior $\bd{\Pi}_i,$ and $u$ is the average of the diagonal entries of $\bd{U_2}.$ 

Simulations and inference with dynamic parameters also require a description of the dynamics. The covariance $\bd{\Omega}_i$ of the Gaussian noise that drives the dynamics for entity $i$ was obtained similarly to $\bd{\Pi}_i$, but with a few differences to achieve both positive and negative covariance entries. The entries of $\bd{U_1}$ are now sampled from a standard normal distribution, then $\bd{U_2} = \bd{U_1}\bd{U_1'}/k_i^2$ as before, and finally  $\bd{\Omega}_i =\frac{s_d}{u} \bd{U_2},$ where $s_d$ is a typical value of a parameter drift per observation we want to achieve, and where the normalizer $u$ is given by $(\det{\bd{U_2}})^{\frac{1}{k_i}}.$ The memory parameter $\alpha_i$ for entity $i$ is set by first choosing the desired half life $t_{hi}$ of the entity, i.e., the number of observations after which any difference from the reference vector should decay in half assuming simple geometric decay. Then, $\alpha_i=\exp{\big(\log{(0.5)} / t_{hi}\big)}$. The rest of the parameters for the different models we study are described next.
\begin{enumerate}
\item \textbf{Regression.} 
We considered a `sparse' regression model roughly equivalent to matrix factorization with known item vectors. We created $10$ entities corresponding to $10$ users with $k_i=30,$ and prior mean $\pi_i$ with all entries equal to $-0.00405.$ In each observation only one of these entities was randomly selected.  We also created an additional entity with $50$ entries that appears in every observation and prior mean with all entries equal to $-0.0068$, to model purely item-dependent effects (like the genre of an item); purely user-dependent or other context-dependent effects like time of day or day of week can be modeled similarly. In each observation, the signal is the dot product of a context vector with $80$ entries and a vector that concatenates the two entities in the observation. So equivalently, this can be thought of as a regression model with $350=10\times30+50$ parameters, and where the context vector has one `dense' portion of $50$ entries that is typically non-zero in all observations, and $10$ `sparse' portions with $30$ entries each, only one of which is non-zero in each observation. All entities used the same dynamics parameters: the half life was set to $10000$, and the covariance scales were set to $s_p=0.005$ and $s_d = 0.0028.$  A fixed set of $100$ context vectors with $80$ entries each were constructed by sampling each entry independently from $N(1, 1)$. In each observation, one of these $100$ context vectors is randomly selected, and combined with the $80$-dimensional parameter vector resulting from selecting one sparse entity and the dense entity, to produce the signal $\lambda$ for the observation.

\item \textbf{Matrix factorization.} We generated $10$ users and $10$ item entities, each with $10$ entries, for a total of $200$ parameters. We set the prior mean entries of $\pi_i$ to $0.2$ for users entities, and to $-0.2$ for item entities (perturbed very slightly per simulation). The half life of all user and item entities was set to $10000,$  and their covariance scales to $s_p=0.144$ and $s_d=$\num{2.45e-5.}.

\item \textbf{Tensor factorization} We decomposed a multi-way array with four modes with dimensions $[3,3,4,4]$, so the number of entities is $14$.  We used a rank $20$ decomposition, so each entity vector was in $\RR^{20}$. We set the entries of $\pi_i$ for all entities to $-0.405465$ (with small random perturbations per simulation). We set the halflife of all entities to $10000,$ and the covariance scales to $s_p=0.23$ and $s_d=$\num{3.8e-5.}.
\end{enumerate}

\subsection{Prediction on Simulated Data}
\begin{figure}
    \centering
    \begin{subfigure}[t]{0.45\textwidth}
        \centering
        \includegraphics[width=\linewidth]{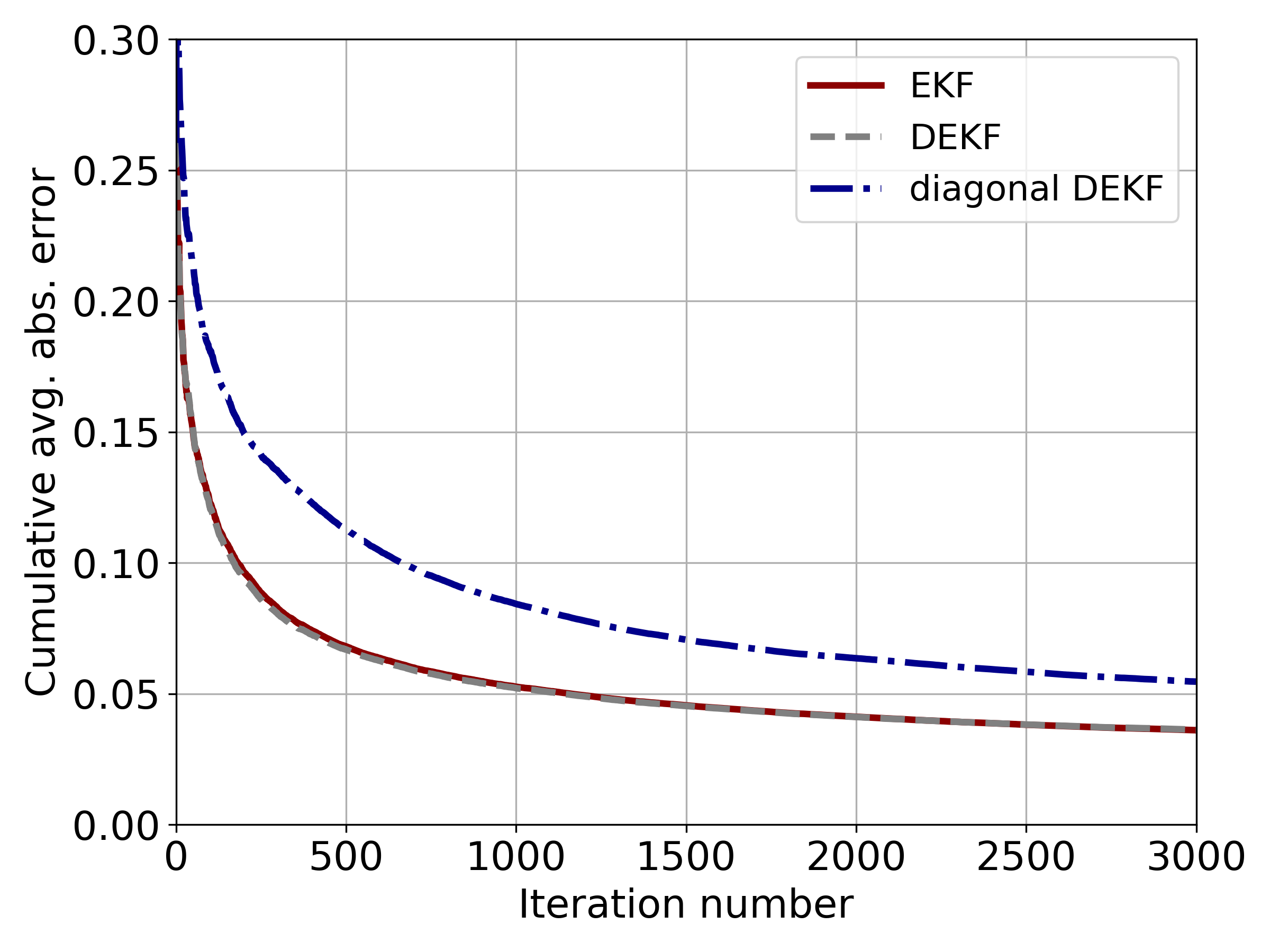}
        \caption{Static regression} \label{fig:reg_stat}
    \end{subfigure}
    \begin{subfigure}[t]{0.45\textwidth}
        \centering
        \includegraphics[width=\linewidth]{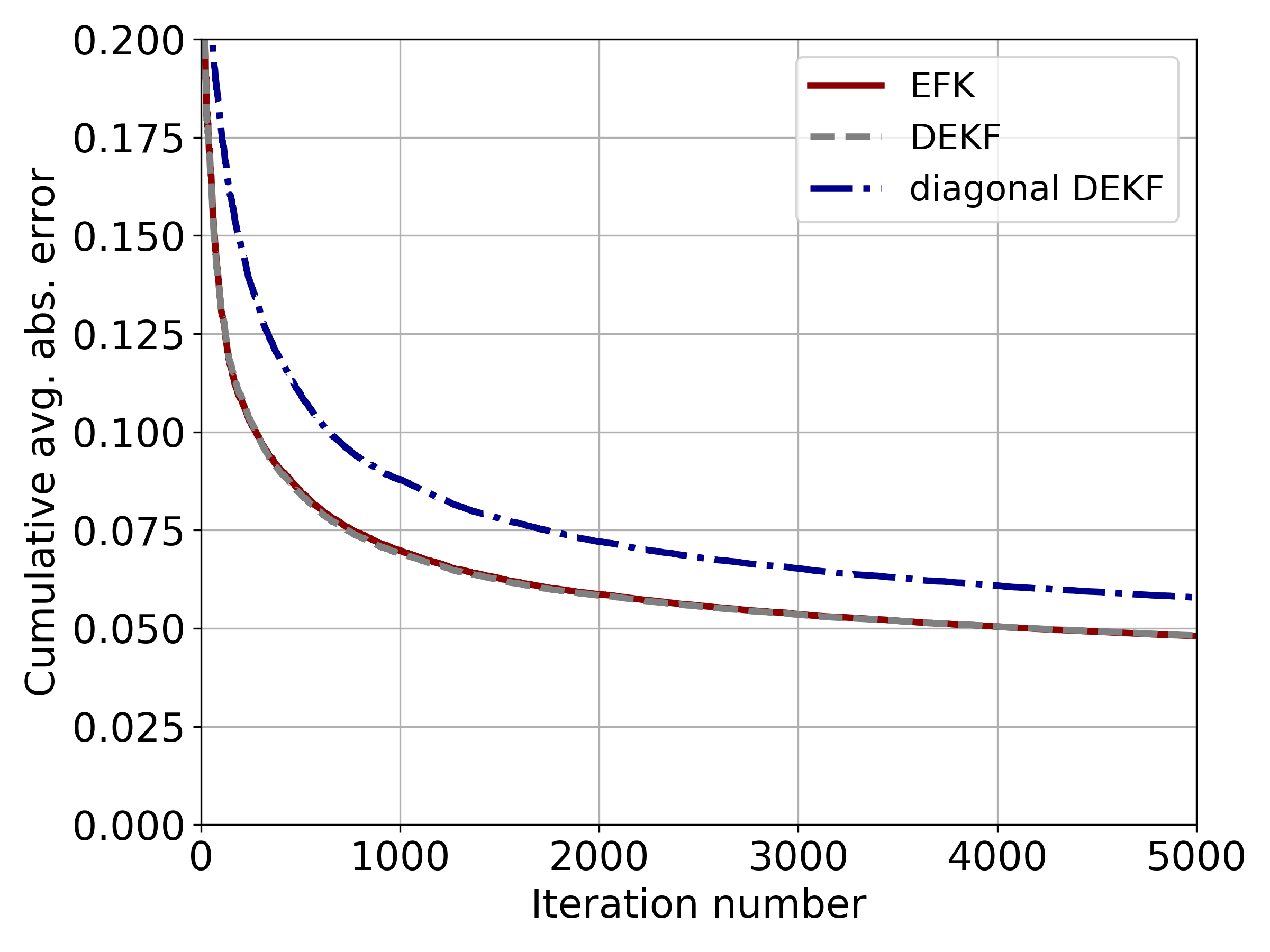}
        \caption{Dynamic regression} \label{fig:reg_dyn}
    \end{subfigure}
    \begin{subfigure}[t]{0.45\textwidth}
        \centering
        \includegraphics[width=\linewidth]{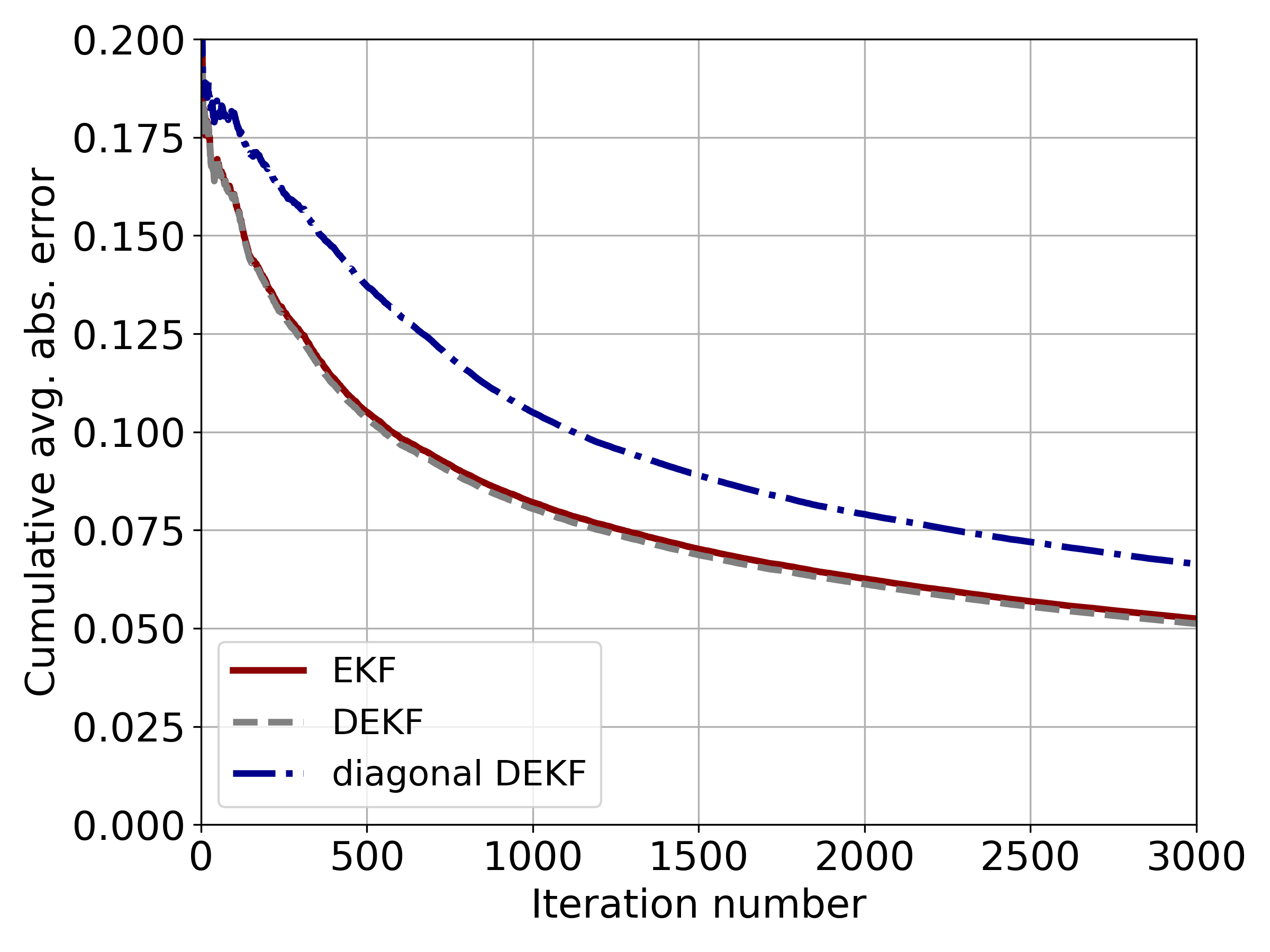}
        \caption{Static matrix factorization} \label{fig:mf_stat}
    \end{subfigure}
    \begin{subfigure}[t]{0.45\textwidth}
        \centering
        \includegraphics[width=\linewidth]{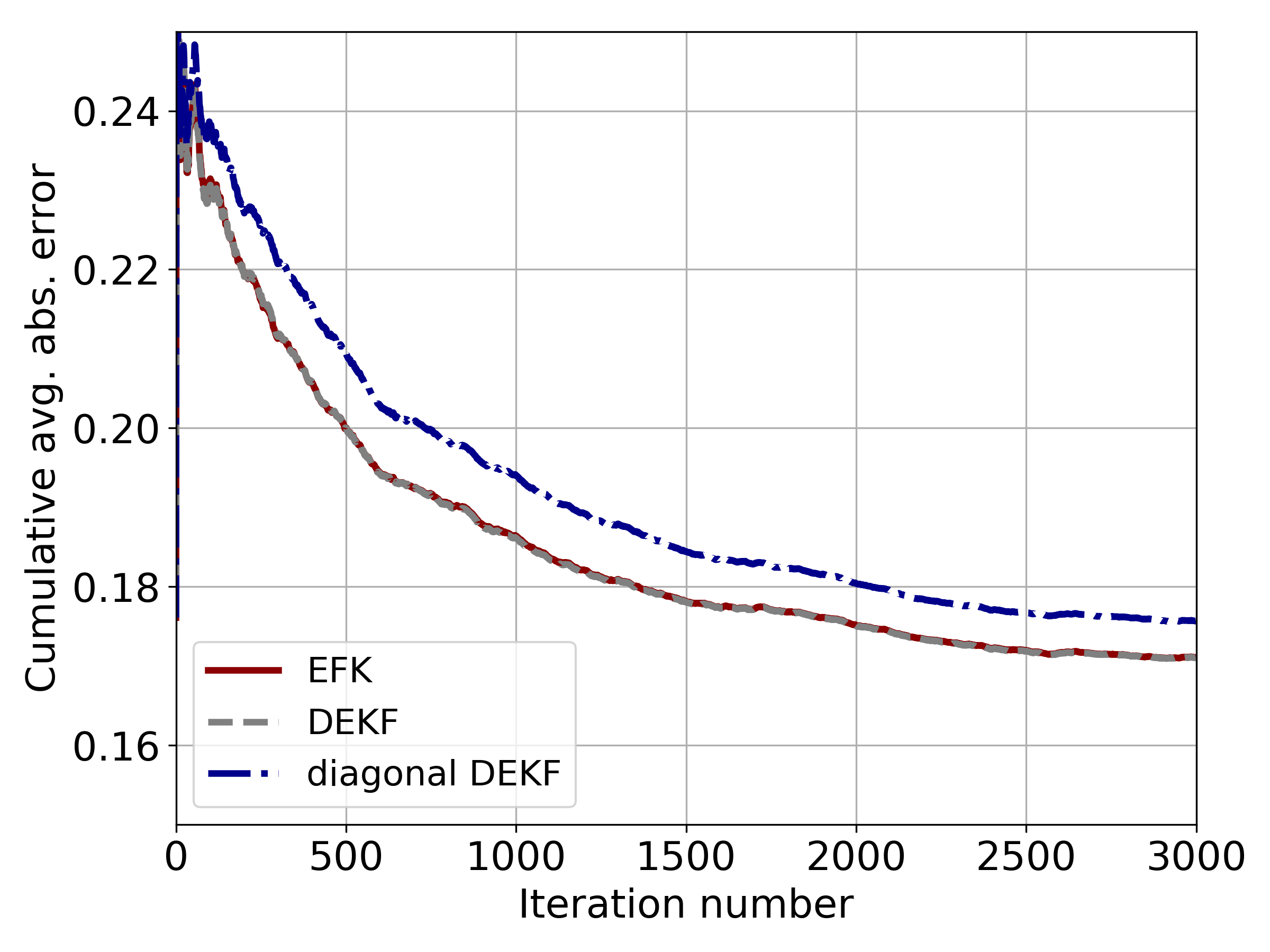}
        \caption{Dynamic matrix factorization} \label{fig:mf_dyn}
    \end{subfigure}
    \begin{subfigure}[t]{0.45\textwidth}
        \centering
        \includegraphics[width=\linewidth]{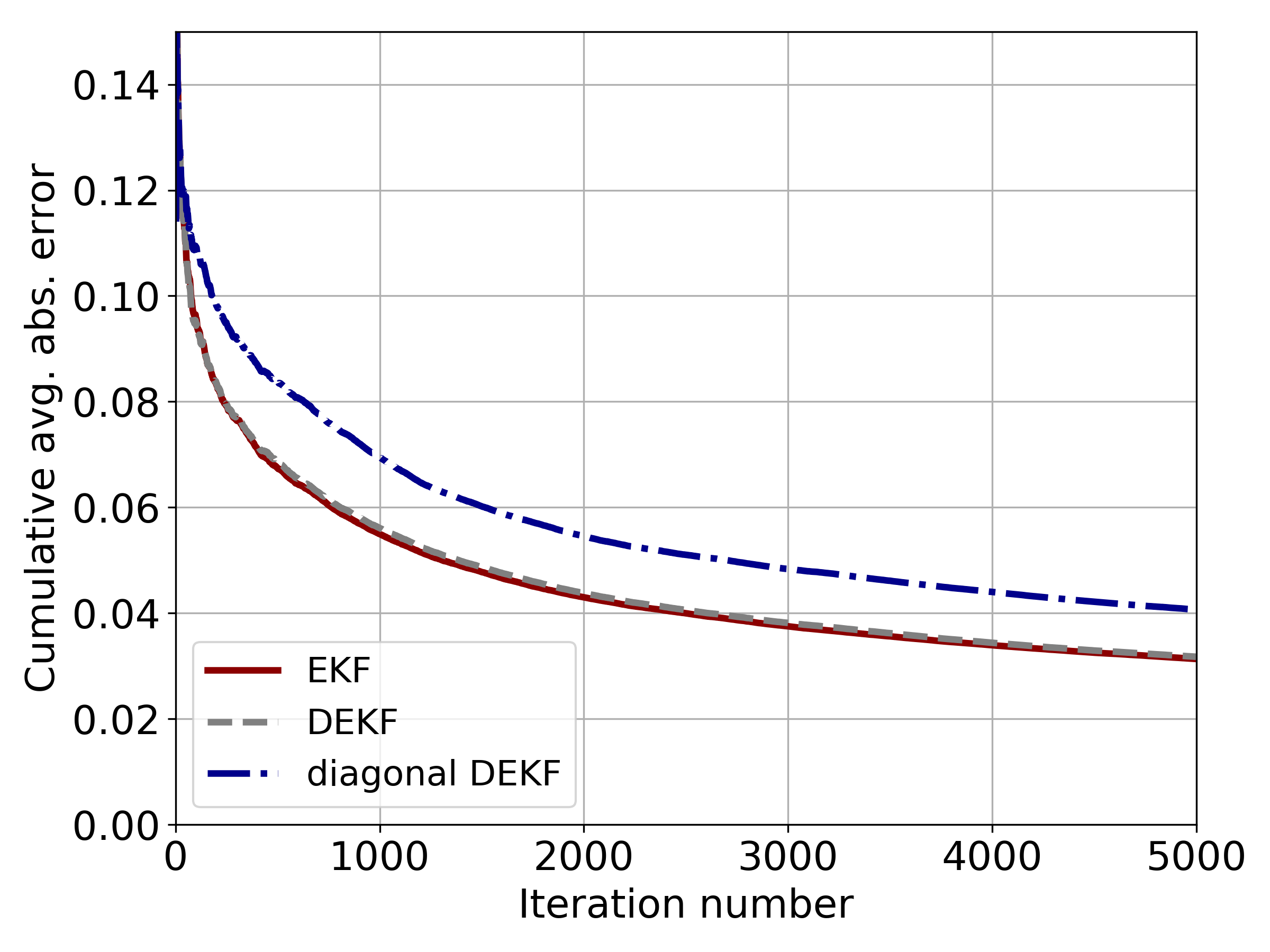}
        \caption{Static tensor factorization} \label{fig:tf_stat}
    \end{subfigure}
    \begin{subfigure}[t]{0.45\textwidth}
        \centering
        \includegraphics[width=\linewidth]{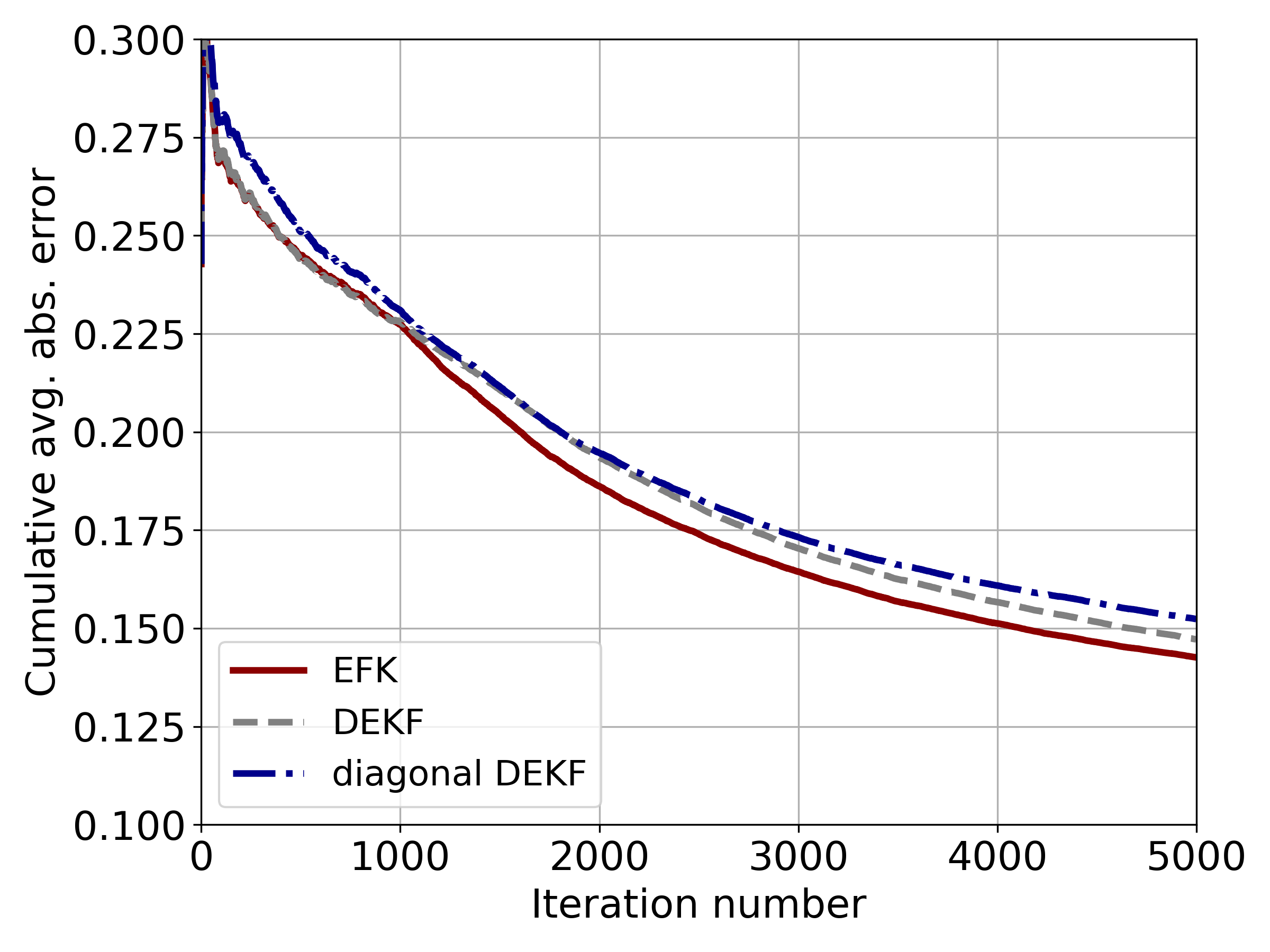}
        \caption{Dynamic tensor factorization} \label{fig:tf_dyn}
    \end{subfigure}
    \caption{Parameter estimation.  The solid lines show the cumulative average absolute error at iteration $t$: $1/t \sum_{i=1}^{t} |p_{true,i} - p_{predicted,i}|$.  
    There is one observation per iteration. All lines are averages over 10 simulations.}
    \label{fig:estimation}
\end{figure}

We track estimation quality by recording how our prediction $h(\mu)$ tracks the true underlying probabilities used to generate the data.\footnote{The true mean prediction averaged over the prior is $E_\theta[h(\theta)]$ but this is impractical to use in general, so we utilize $h(\mu)$ in our simulations, in close correspondence with our update equations.} 
We first consider sparse regression, MF and TF with static parameters estimated via Algorithm~\ref{algo:framework}. The results for prediction are shown on the left-hand column of Figure~\ref{fig:estimation}. The DEKF performance is practically identical to that of the full EKF on this scale for the three models, showing that the block diagonal covariance approximation of the DEKF with one block per entity is adequate. A full diagonal approximation of the covariance, labeled diagonal DEKF in our plots, performs decently but worse that the EKF and the DEKF.

When we allow for the specified parameter dynamics, and estimate them via Algorithm~\ref{algo:frameworkWithDrift}, which takes these dynamics into account, we obtain the results in the right-hand side of Figure~\ref{fig:estimation}. Again, we see that the EKF and DEKF perform similarly, and better than the diagonal DEKF. In dynamic TF, Figure \ref{fig:tf_dyn}, we actually see the full EKF perform a little better than the DEKF. This is perhaps expected since TF is a more non-linear function of the parameters than MF and regression, resulting in a covariance of the parameters farther from block diagonal. As expected, we also find (not shown) that assuming static parameters for inference when there are parameter dynamics, i.e., using Algorithm~\ref{algo:framework} for models with dynamic parameters, results in significantly worse performance than using the inference that accounts for parameter dynamics. Of course, on real data sets, one does not know the true parameters for the dynamics, and mis-specification of these can lead to degraded performance.

\subsection{Explore-Exploit on Simulated Data}
The posterior uncertainty can be utilized for exploration and exploitation, regardless of dynamics in parameters. To demonstrate this, we use the same setup as above, except that at each time step $t,$ where a user is randomly selected, we must \textit{recommend} the remaining entities or context to completely specify the observation. The goal is maximizing the number of observations with a positive outcome. In sparse regression, we randomly select one sparse entity (the user) per observation, and then choose one of the $100$ context vectors (an item) as the recommendation. In matrix factorization, a specific user is sampled at time $t,$ and we recommend an item entity to that user. In our TF model, since it has four kinds of entities, once a specific user is sampled at time $t,$ we need to choose the three entities of the three remaining types to define the observation. Once a recommendation has been made, the corresponding observation is sampled and the outcome recorded. To perform the recommendation, the algorithm begins by applying the predict step for the means and covariances of every entity in the set of contexts.\footnote{In a practical application, this predict-sample cycle would not need to be applied on every recommendation.}  We then generate a recommendation through Thompson sampling \citep[see][for an overview]{russo2017tutorial}\footnote{We generate one posterior sample $\theta_{sampled}$ from the joint posterior over every entity possibly in the observation, and then select a valid set of sampled entities and context $x$ to maximize $h(\theta_{sampled}, x)$.} and apply the update step after receiving a new observation $y_t$ for the recommendation from the underlying process.  

We evaluate recommendation quality by measuring cumulative regret, the sum over recommendations of the true probability for the best context minus the true probability for the chosen context.  We compare Thompson sampling against random recommendations as well as recommending the context with the highest prediction $h(\mu)$, an approach that does not require posterior uncertainty.  E.g., for traditional MF this strategy recommends the item vector with the largest dot-product with the user vector. To reduce the computational requirements, we changed some of the parameters of our models. Specifically, for sparse regression, we now let the sparse entities have only $10$ entries, and the dense one $20.$ For TF, we reduced the number of entries per entity to $5.$ For MF, we slowed down parameter dynamics by setting the half life of all entities to $100000$.

\begin{figure}
    \centering
    \begin{subfigure}[t]{0.45\textwidth}
        \centering
        \includegraphics[width=\linewidth]{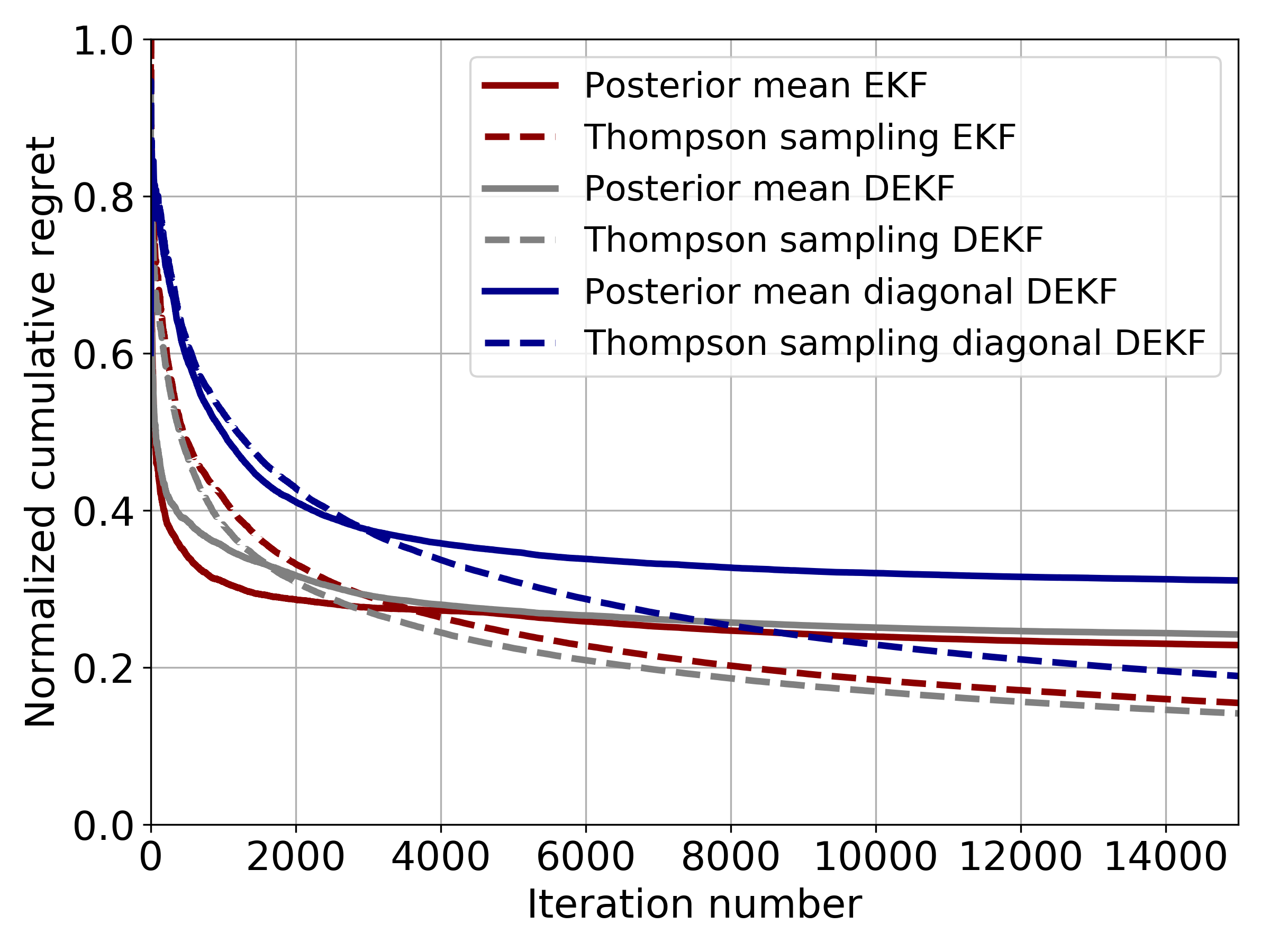}
        \caption{Static regression} \label{fig:reg_explore_static}
    \end{subfigure}
    \begin{subfigure}[t]{0.45\textwidth}
        \centering
        \includegraphics[width=\linewidth]{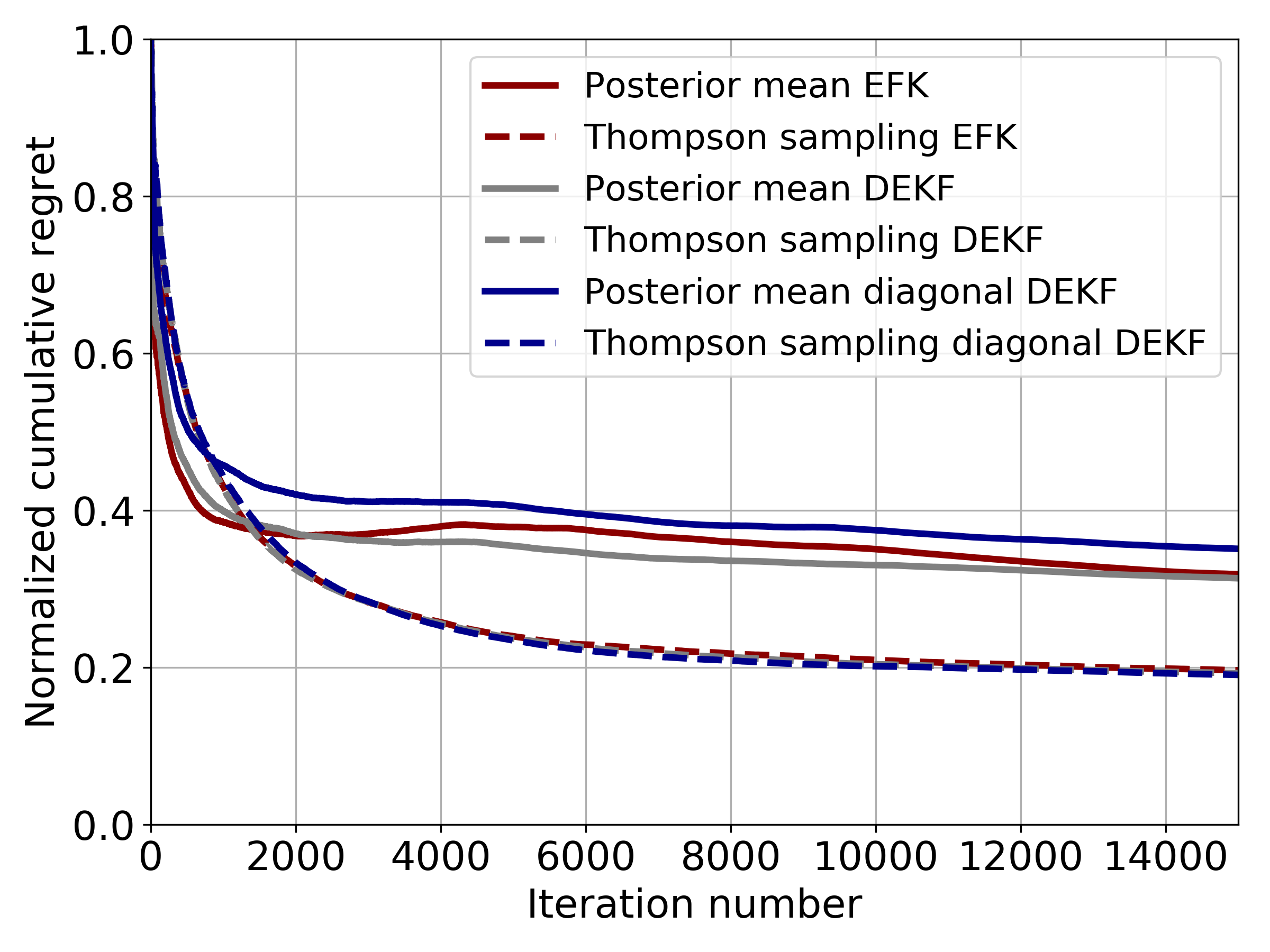}
        \caption{Dynamic regression} \label{fig:reg_explore_dynamic}
    \end{subfigure}
    \begin{subfigure}[t]{0.45\textwidth}
        \centering
        \includegraphics[width=\linewidth]{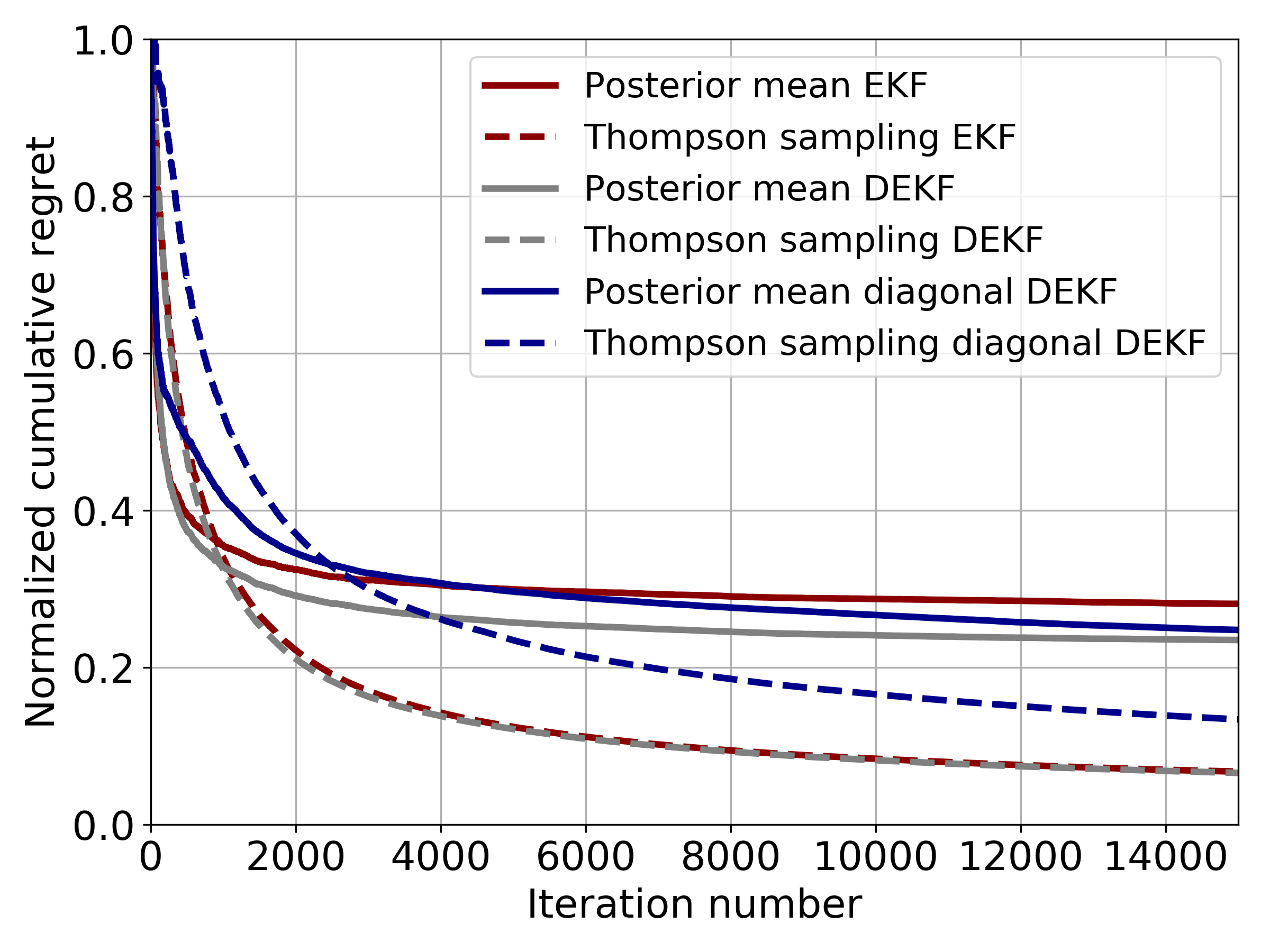}
        \caption{Static matrix factorization} \label{fig:mf_explore_static}
    \end{subfigure}
    \begin{subfigure}[t]{0.45\textwidth}
        \centering
        \includegraphics[width=\linewidth]{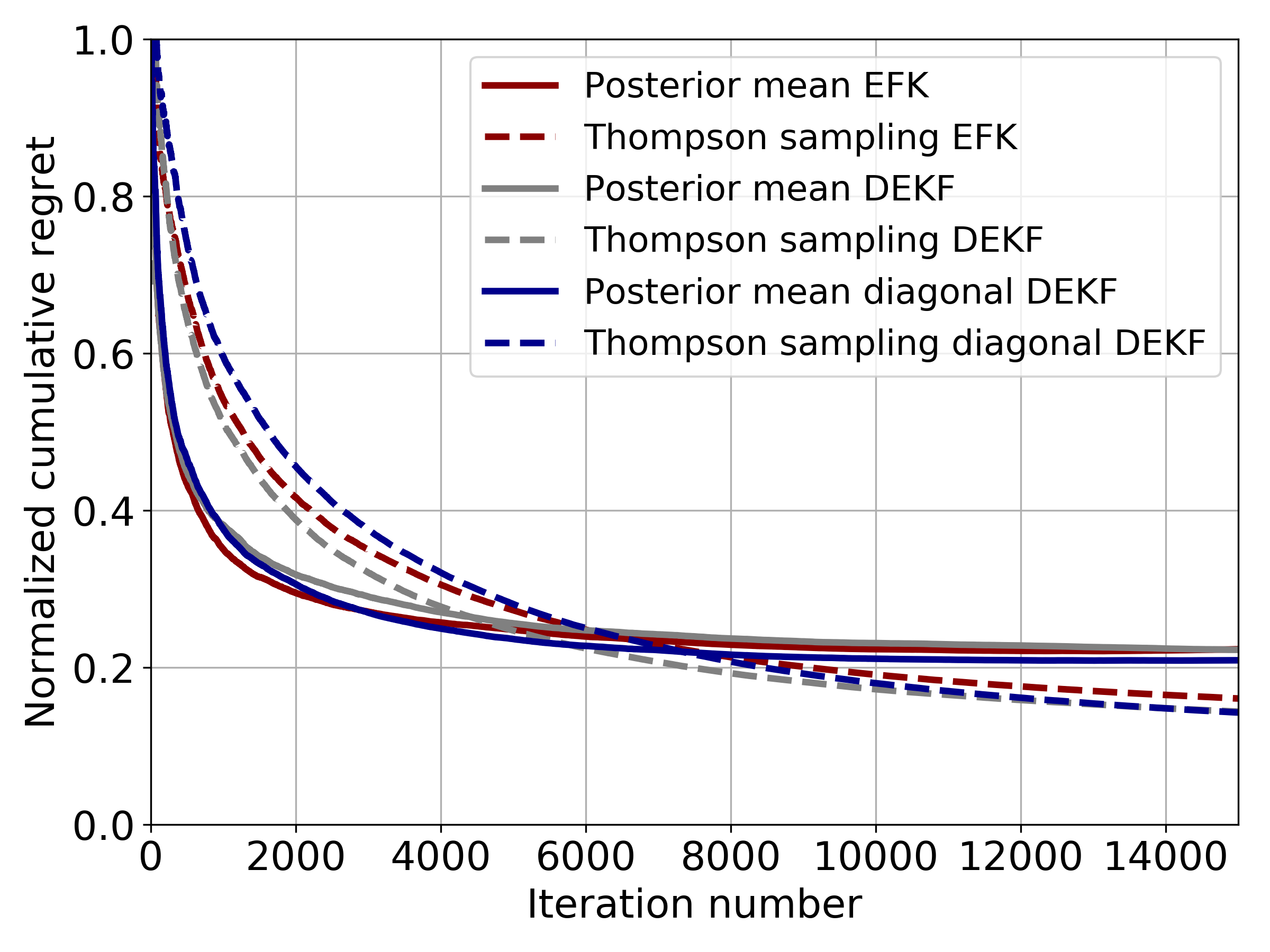}
        \caption{Dynamic matrix factorization} \label{fig:mf_explore_dynamic}
    \end{subfigure}
    \begin{subfigure}[t]{0.45\textwidth}
        \centering
        \includegraphics[width=\linewidth]{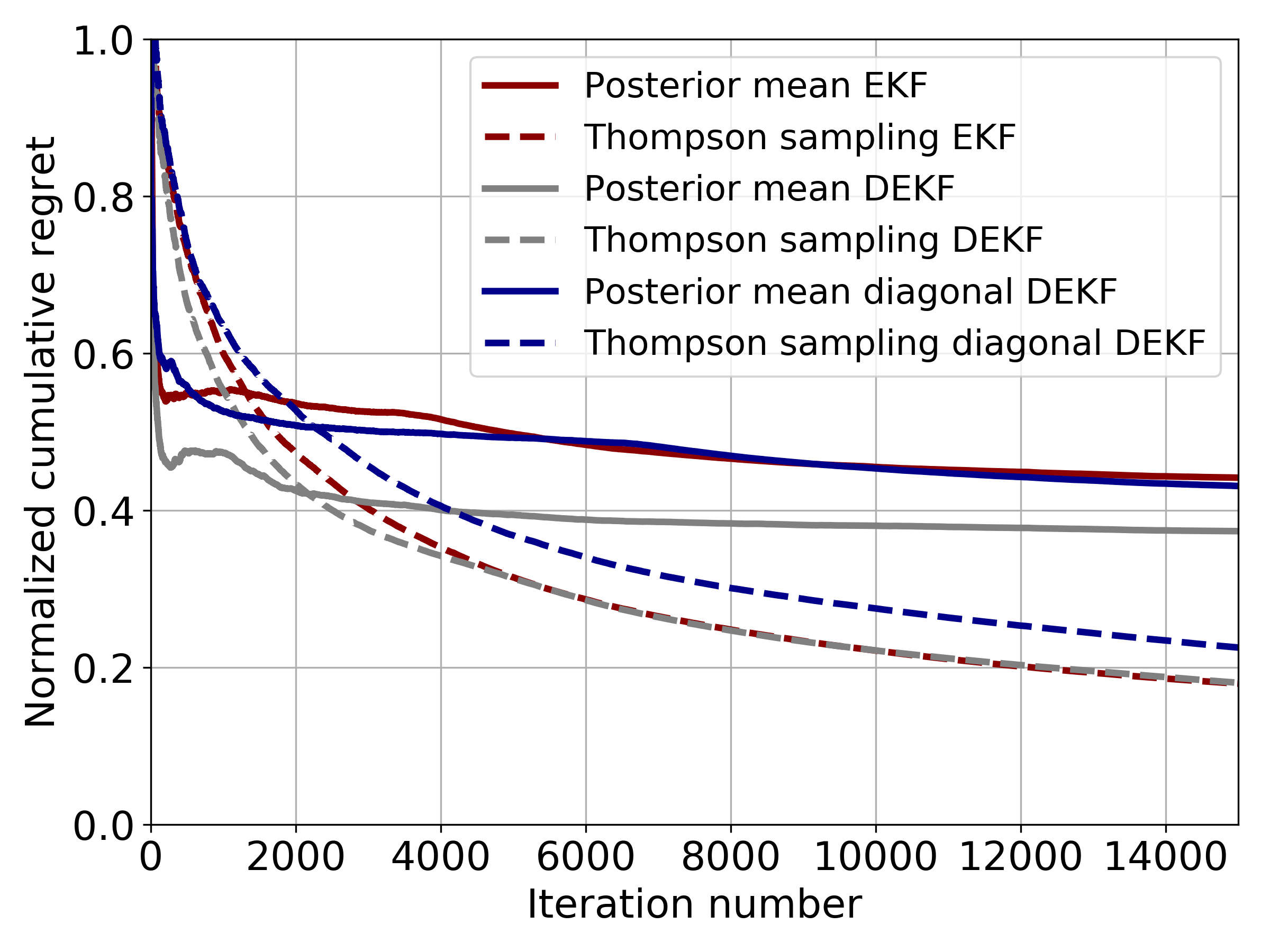}
        \caption{Static tensor factorization} \label{fig:tf_explore_static}
    \end{subfigure}
    \begin{subfigure}[t]{0.45\textwidth}
        \centering
        \includegraphics[width=\linewidth]{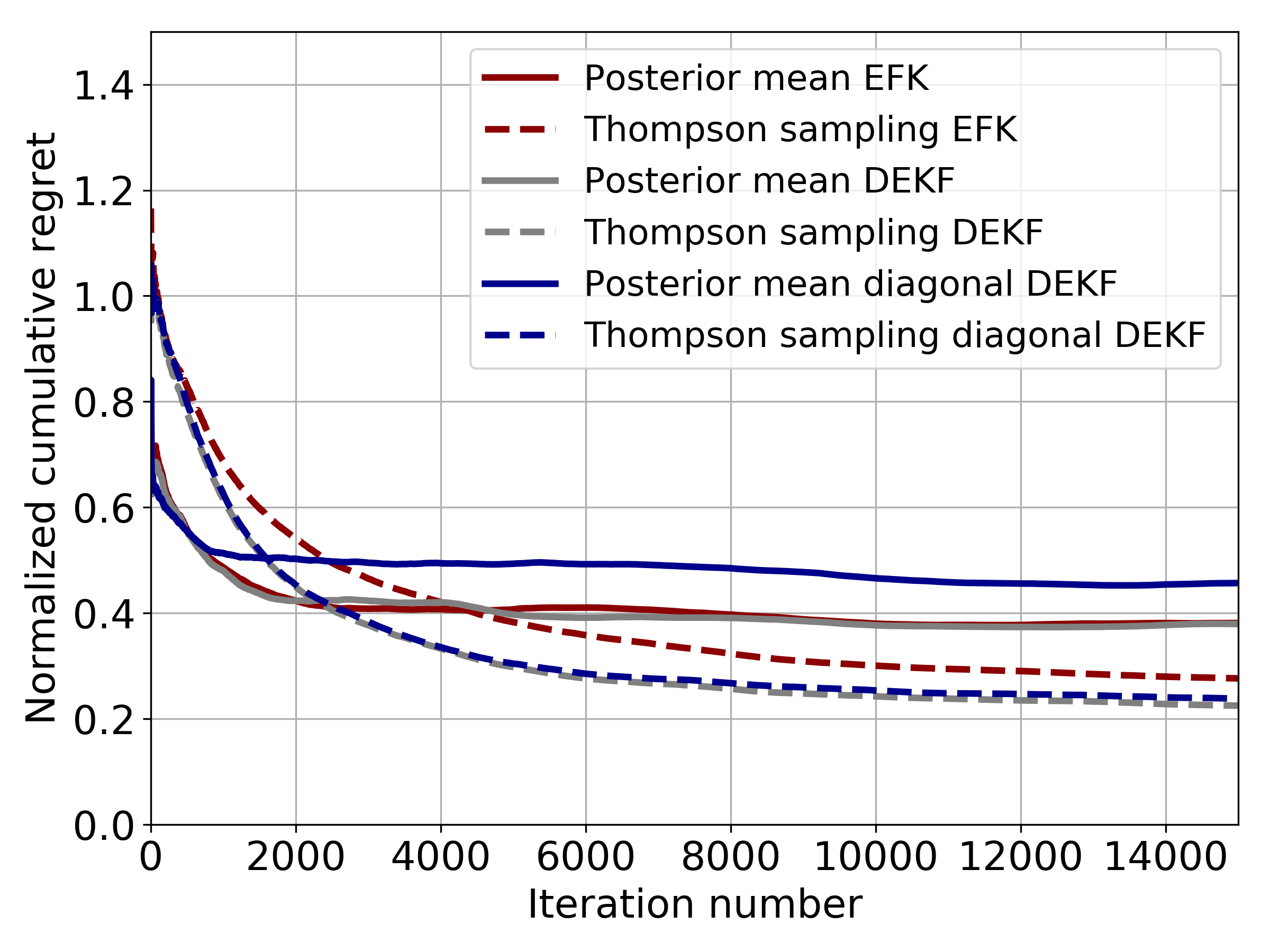}
        \caption{Dynamic tensor factorization} \label{fig:tf_explore_dynamic}
    \end{subfigure}
    \caption{Explore-exploit.  Each line is the normalized cumulative regret at iteration $t$, defined as $\sum_{i=1}^{t}(p_i - q_i)$, divided by the same quantity for random recommendations, where $p_i$ and $q_i$ are the highest probability context and the probability of the recommended context at time $i$ respectively.  Figures show averages over 20 simulations.}
    \label{fig:explore_exploit}
\end{figure}

\begin{figure}
    \centering
    \begin{subfigure}[t]{0.45\textwidth}
        \centering
        \includegraphics[width=\linewidth]{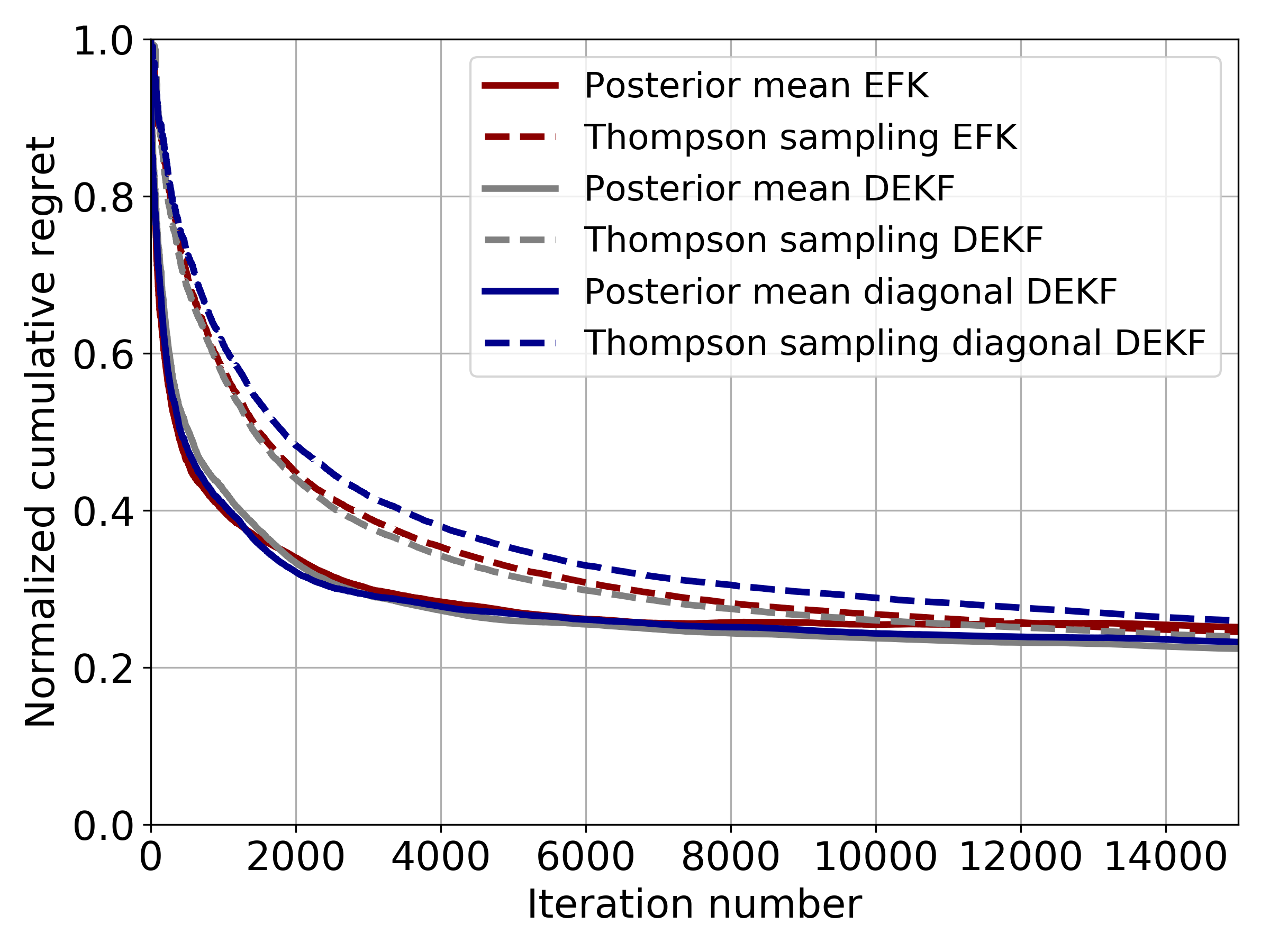}
        \caption{Dynamic matrix factorization with half life of 10000.} \label{fig:mf_explore_dynamic_10000}
    \end{subfigure}
    \begin{subfigure}[t]{0.45\textwidth}
        \centering
        \includegraphics[width=\linewidth]{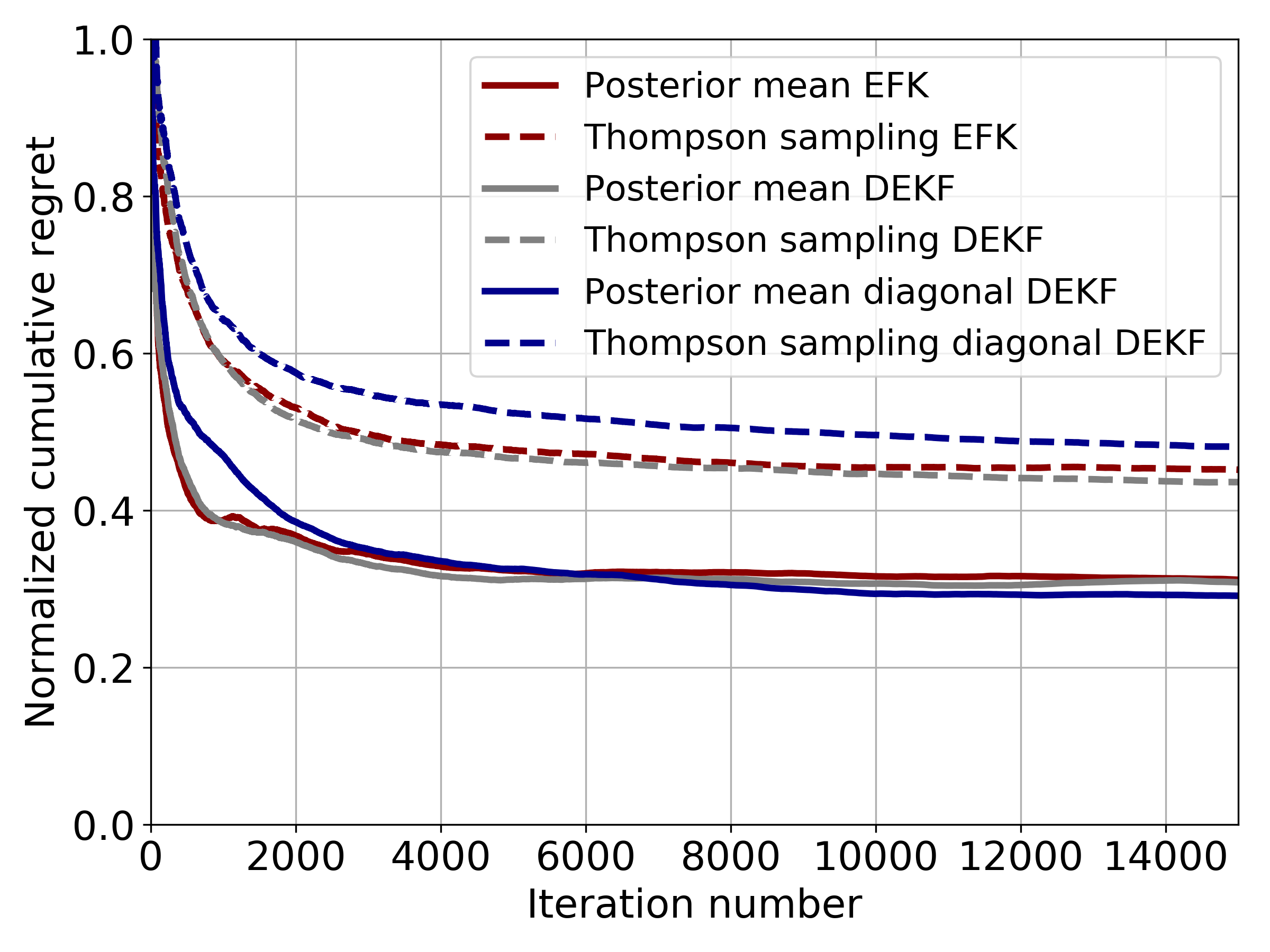}
        \caption{Dynamic matrix factorization with half life of 1000.} \label{fig:mf_explore_dynamic_1000}
    \end{subfigure}
    \caption{Explore-exploit with stronger parameter drift. Dynamic MF with half lives of $10000$ and $1000$, instead of $100000$.}
    \label{fig:explore_exploit_MF}
\end{figure}

The left-hand and right-hand side of Figure~\ref{fig:explore_exploit} respectively show the cumulative regret for static and dynamic parameters, normalized through division by the cumulative regret achieved by random recommendations. We see that leveraging the uncertainty through Thompson sampling eventually results in a substantially lower cumulative regret than using the (approximate) posterior mean for static parameters, in all cases. We also see that the EKF and DEKF perform similarly when using Thompson sampling, while the diagonal DEKF performs worse.  For dynamic parameters, we do not see much difference in the performance between the diagonal DEKF, the DEKF, and the EKF. Thompson sampling eventually performs better than recommending based on the posterior mean in these examples, just like with static parameters. However, such behavior is not general, and in fact, the relative performance of Thompson sampling and posterior mean recommendation depends strongly on the balance between the average information obtained from an observation and the average information lost because of parameter drift. Thompson sampling can be inefficient because of over-exploration in situations when the system changes over time faster than the observations provide useful information to determine the optimal action, e.g., see section 8.2 in  \citet{russo2017tutorial}. As a simple example, Figure~\ref{fig:explore_exploit_MF} shows what happens when we increase parameter drift in our MF model by reducing the half life to $10000,$ and then further to $1000.$ The performance of Thompson sampling worsens indeed, eventually making recommendations based on the posterior mean a better strategy.

\subsection{Prediction on Real Data}

We now apply the DEKF to two data sets, MovieLens-20M~\citep{Harper:2015}) and NetflixPrize~\citep{Bennett07thenetflix}).  MovieLens-20M contains around 20 million ratings of 27k movies by 138.5k users from January 1995 to March 2015.  The ratings were one to five stars until February 2003, at which point they include half-star ratings.  The MovieLens platform that collects this data set has undergone numerous changes, and historical details can be found in~\citet{Harper:2015}.  The NetflixPrize data set contains roughly 100 million ratings of 17k movies by 480k users from October 1998 and December 2005.  Each rating is on a five star scale like early MovieLens.

Both data sets are commonly used as benchmarks for evaluating factorization models in a supervised learning setting, through test-train splits. Such an approach can make sense to evaluate batch algorithms, but we address a different problem: online learning where the underlying data distribution changes over time. So we mimic online learning on these data sets by ordering and analyzing the data points chronologically. The resulting metrics are then not directly comparable to other batch approaches, e.g., that have the benefit of having learned from data points across the entire time range of the data sets.

Because most user-item pairs do not have ratings, we focus on prediction, and not explore-exploit. Both data sets are too large for the EKF in models where each user and each item require their own parameters. But the DEKF still allows us to learn a matrix factorization model where each user and each item is an entity. Each user and movie vector is chosen to be ten-dimensional. We model ratings in two ways, resulting in two different models: as observations from a Gaussian distribution (Gaussian-MF) with standard deviation chosen to be a quarter-star, and as Bernoulli observations (Bernoulli-MF) corresponding to whether the star rating was greater than or equal to four.    

We consider four DEKF versions, depending on whether we treat each parameter entry as an entity (i.e., a diagonal DEKF) or each user and item vectors as an entity (i.e., the standard DEKF), and on whether we consider the parameters to be static or dynamic.  We initialize each variant to have the same priors, utilizing the mean and standard deviation of the observations to set prior mean $\pi$ and prior covariance scale $s_p$.  For dynamics, we use half-lives of one year and five years for users and movies, respectively, for MovieLens-20M, and one year for both users and movies for NetflixPrize.  In contrast to our study based on simulated data, here we set $\bd{\Omega}$ to be diagonal with identical entries equal to $s_d$, allowing for different values for users and items.\footnote{We tried several values for these dynamic parameters, and chose the particular dynamic parameters via examining predictions on the first five thousand observations, a minute fraction of the observations.}

\subsubsection{MovieLens-20M}

In Figure~\ref{fig:movielens_gaussian}, the cumulative average out-of-sample root-mean-square-error (RMSE) is shown as a function of time for Gaussian observations.  Vertical blue lines denote dates for significant changes to the MovieLens platform from Table $1$ in~\citet{Harper:2015}.  The effects of these changes are clearly visible in the algorithm's curves.  In particular, the switch to include half-star ratings in February 2003 substantially lowered RMSE.  In this figure, the diagonal DEKF performs worst, and modeling parameter dynamics was helpful.  To interpret these RMSE values, we reference Table $1$ from~\citet{StrubMG16a} where the best supervised learning MovieLens-20M result had a batch RMSE evaluated on a test set of $0.7652$, and a rank $10$ Bayesian probabilistic matrix factorization (BPMF) achieved $0.8123$.  Differences between approaches were on the order of hundredths of RMSE, indicating that the observed differences between DEKF versions is substantial.  Our best performing algorithm achieves RMSE of $0.8082$, comparable to the BPMF, despite our approach and RMSE evaluation being online, as already stressed before.  
In Figure~\ref{fig:movielens_bernoulli}, for the Bernoulli observations, we show the cumulative average out-of-sample normalized cross-entropy (NE).\footnote{Let $L(y, p) = -y \log(p) - (1-y) \log(1-p)$.  Let $p_i$ and $y_i$ be the prediction and value of the $i$th observation respectively, and $p_{base} = 1/n  \sum_{i=1}^{n} y_i$.  Cumulative average NE at iteration $t$ is $\sum_{i=1}^{t} L(y_i, p_i) / \sum_{i=1}^{t} L(y_i, p_{base})$.}  Blue lines again indicate substantial platform changes.  The effects of platform changes are still visible, although less so for the Bernoulli observations.  Dynamics is again helpful, although surprisingly, here we observe the diagonal DEKF performs best.  

\subsubsection{NetflixPrize}

\begin{figure}
    \centering
    \begin{subfigure}[t]{0.45\textwidth}
        \centering
        \includegraphics[width=\linewidth]{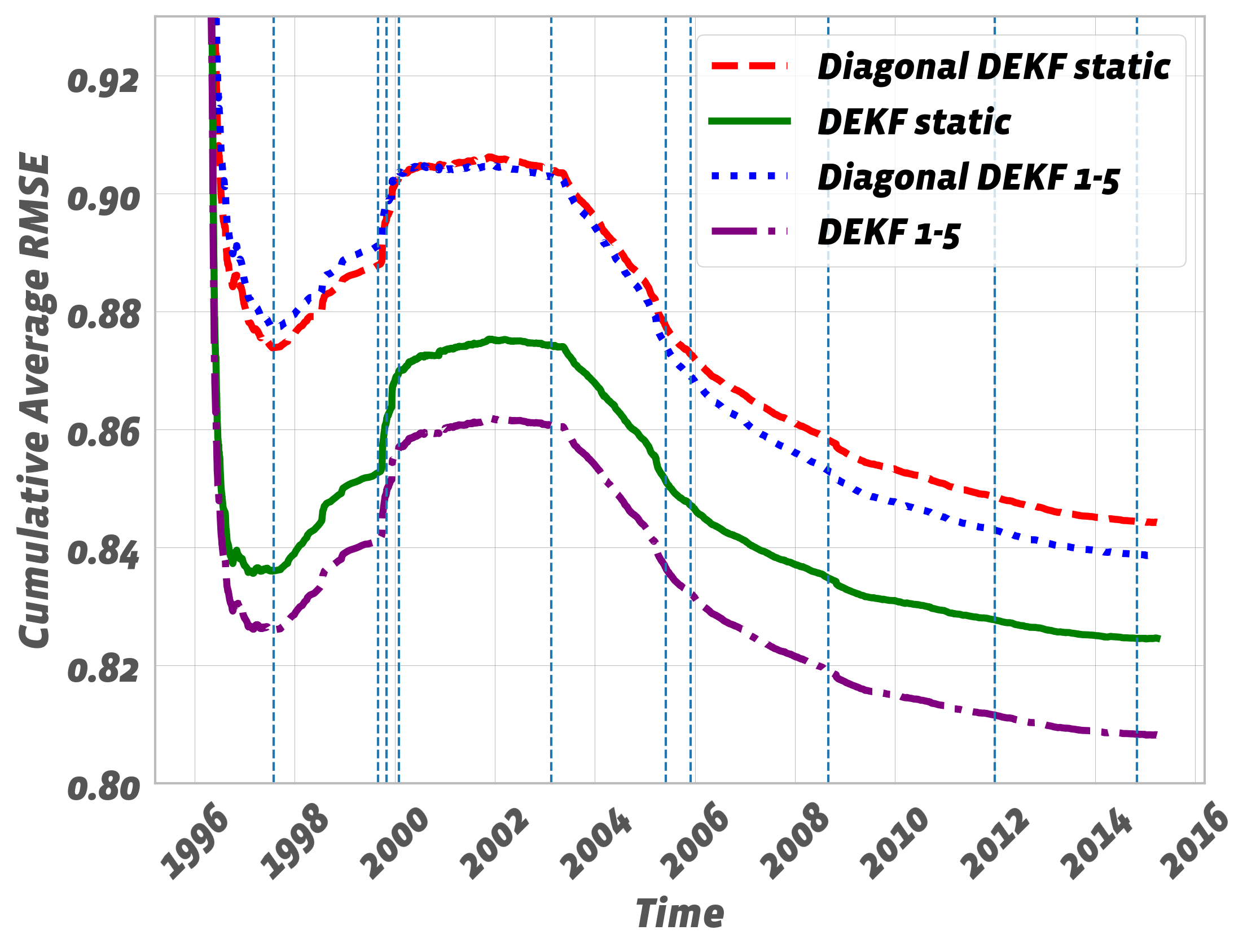}
        \caption{MovieLens-20M Gaussian-MF} \label{fig:movielens_gaussian}
    \end{subfigure}
    \begin{subfigure}[t]{0.45\textwidth}
        \centering
        \includegraphics[width=\linewidth]{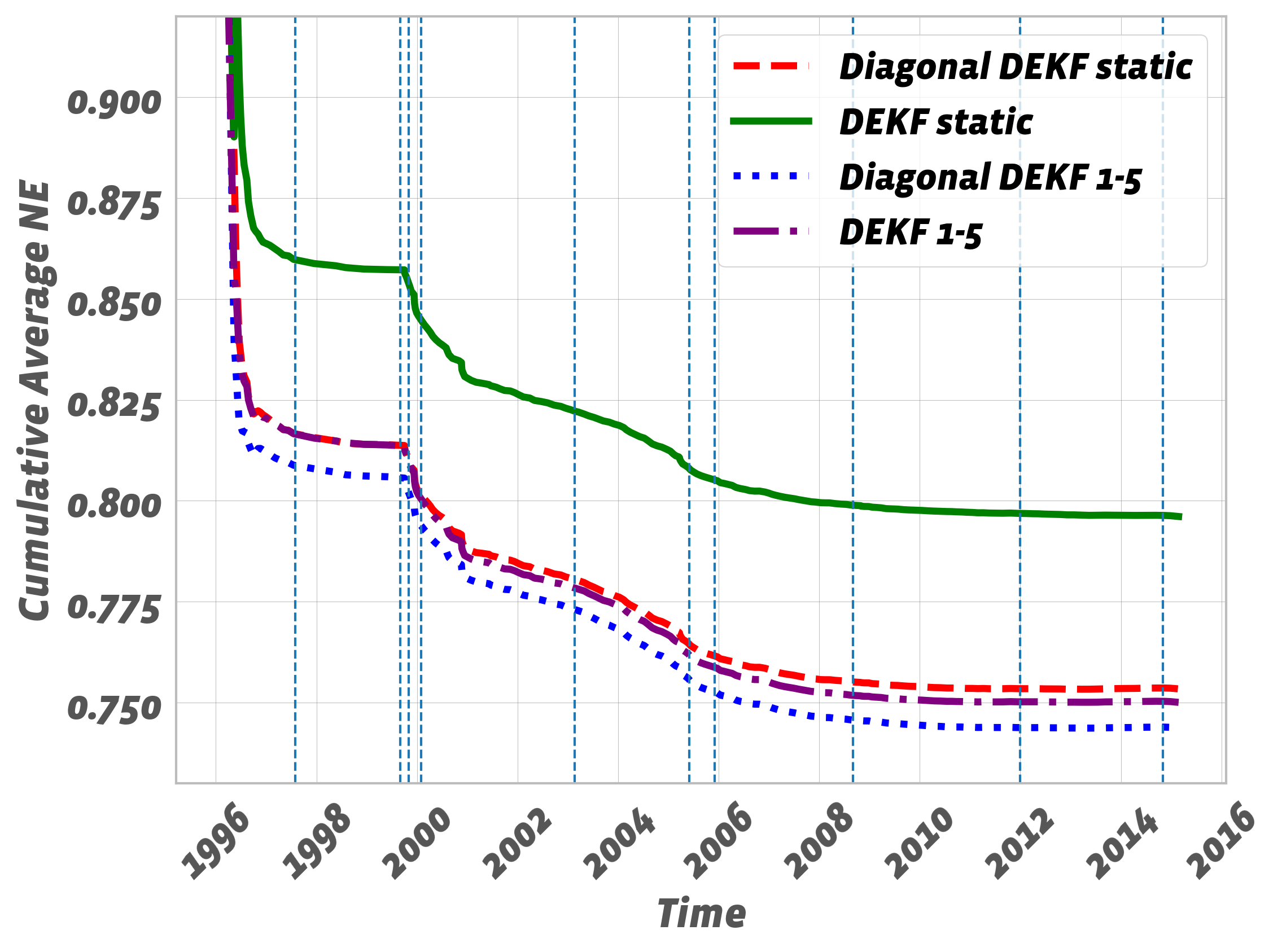}
        \caption{MovieLens-20M Bernoulli-MF} \label{fig:movielens_bernoulli}
    \end{subfigure}
    \begin{subfigure}[t]{0.45\textwidth}
        \centering
        \includegraphics[width=\linewidth]{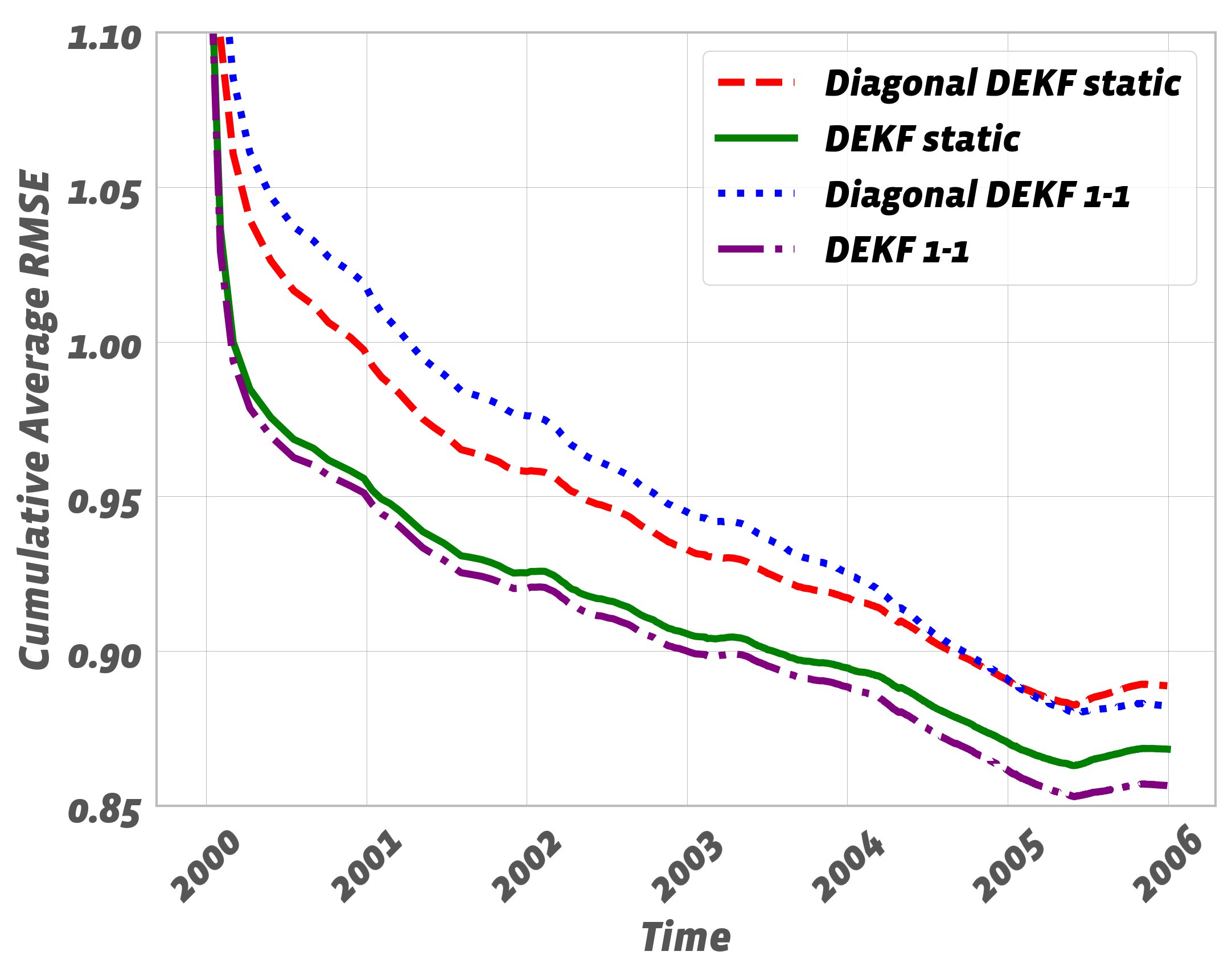}
        \caption{NetflixPrize Gaussian-MF.} \label{fig:netflix_gaussian}
    \end{subfigure}
    \begin{subfigure}[t]{0.45\textwidth}
        \centering
        \includegraphics[width=\linewidth]{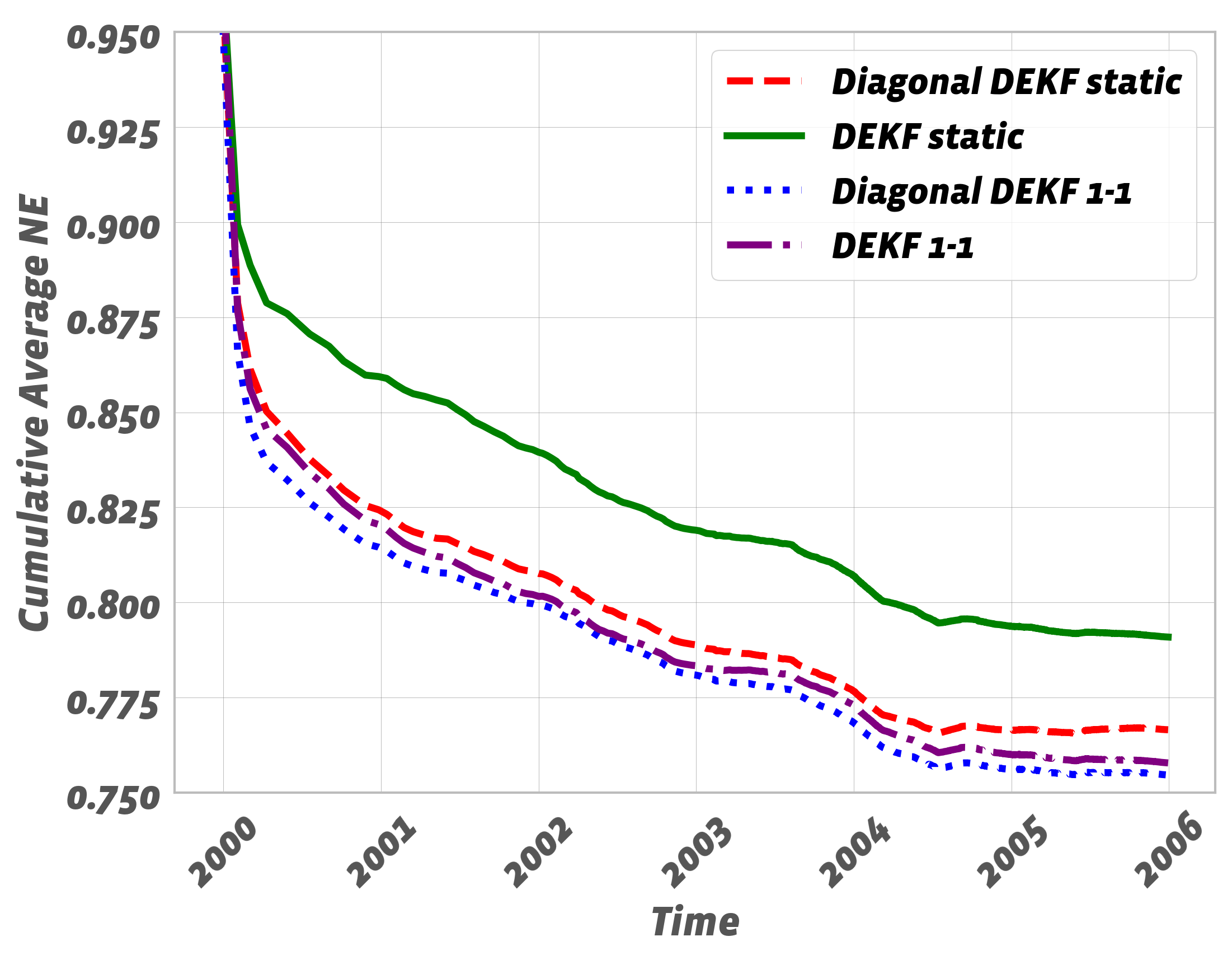}
        \caption{NetflixPrize Bernoulli-MF} \label{fig:netflix_bernoulli}
    \end{subfigure}
    \caption{Static and dynamic DEKF, with diagonal and user / item entity choices, applied to MovieLens-20M and NetflixPrize.  Dynamic half-lives for users and items are included in each legend, and expressed in years. Vertical blue lines in MovieLens-20M correspond to significant changes in that platform detailed in \citet{Harper:2015}. Let $\pi$ be the value of every entry of $\pi_i$ for all entities. For (a), $\pi =0.5916$, $s_p=0.0924$, $s_{d,user}=1.3585\text{e-}9$, and $s_{d,movie}=2.717\text{e-}10$.  For (b), $\pi=4.4721\text{e-}5$, $s_p=0.2133$, $s_{d,user}=7.8633\text{e-}9$, and $s_{d,movie}=1.5727\text{e-}9$.  For (c), $\pi=0.6003$, $s_p=0.0916$, $s_d=2.8279\text{e-}9$.  For (d), $\pi=0.1630$, $s_p=0.1908$, $s_d=6.8433\text{e-}9$.}
    \label{fig:explore_exploit_MF_real}
\end{figure}

We repeat these calculations for NetflixPrize.  In Figure~\ref{fig:netflix_gaussian}, we show the cumulative average out-of-sample RMSE achieved as a function of time for the Gaussian observations.  Similar to MovieLens-20M, the diagonal version performs poorly compared to the DEKF, with a benefit from dynamics.  We compare against supervised learning results, where in Table $3$ from~\citet{ZhengTDZ16}, we observe a range of test RMSE across algorithms from $0.803$ to $0.874$.  Our best performing version has RMSE $0.856$, comfortably within this range, but achieved for online learning. Again for the Bernoulli observations, in Figure~\ref{fig:netflix_bernoulli}, we show the cumulative average out-of-sample normalized cross-entropy (NE).  Like MovieLens-20M, dynamics performed best for Bernoulli-MF, and diagonal entities showed a slight edge.

\subsubsection{Conclusions}

Overall these results indicate that the DEKF produces reasonable predictions on real-world data sets, and that incorporating dynamics can improve predictions.  The best choice of entities for prediction accuracy was inconsistent across the experiments, e.g., setting each parameter as an independent entity was often superior for Bernoulli observations.  This diagonal option is appealing from a complexity standpoint, and for very large data sets, may be the only option available.

\section{Discussion}
\label{sec:discussion}
We have specialized the EKF to a model with observations in the exponential family, which includes the GLM, MF, TF and factorization machines. This treatment results in more flexible observation models than are typically considered in these models. It also enables parameter dynamics to account for data drift. In addition, the uncertainty around the estimates the EKF provides can enable applications where uncertainty is necessary, such as explore/exploit. However, when the number of parameters is large, as is often the case in modern applications, the memory and computation requirements of the EKF can be prohibitive. To address this, we specialize the DEKF to our model. We show that in both the EKF and the DEKF, only parameters involved in an observation need to be updated, and develop an optimized version of the DEKF that is particularly well suited for the kinds of models we consider, which are naturally defined to only involve a relatively small subset of the parameters in each observation.

Of course, the EKF is an approximate inference algorithm and has been observed to sometimes produce badly behaved parameter estimates when the response function is sufficiently non-linear and the initial prior is not sufficiently well-specified.  The DEKF inherits those problems, and examples can be found by applying the DEKF to the Poisson distribution with the canonical link ($h(\eta) = e^{\eta}$), which often displays enormous predictive errors for early iterations.  Fortunately, these problems can have known, simple solutions.  Either one can strengthen the prior, add a learning rate to slow down the initial parameter updates, or utilize the iterated decoupled EKF (IDEKF), as described in Appendix A, instead of the DEKF.  A safe default procedure for highly non-linear responses may be to start with the IDEKF and later switch to the DEKF, but as shown in Section~\ref{sec:numericalResults}, this was unnecessary for our numerical results.  A more serious problem occurs when the true posterior is multi-modal and not well-approximated as a Gaussian.  Like the EKF (and related methods), we expect the DEKF to not perform well in this situation.      

Our approach contains hyperparameters per entity given by $\pi$, $\bd{\Pi}$, $\bd{\alpha}$, and $\bd{\Omega}$. The latter two are only relevant in situations with dynamic parameters, while the first two are always relevant. In specific applications, it is typically unclear a priori whether including dynamics (through $\bd{\alpha} \neq 1.0$, and $\bd{\Omega} \neq \bd{0}$) will result in more useful models.  
Indeed, our simulations suggest that the model choices that better match the true data generation process, which is typically unknown, work best.  On the MovieLens-20M and NetflixPrize data sets, however, we observed that adding dynamics with reasonable settings was helpful.
One way to specify $\pi_i$ and $\bd{\Pi}_i$ would be to analyze offline data about the entity.  An easier approach is to first specify the prior per entity type (e.g., all items are given the same prior).  Then, we recommend sampling entities from these priors (and possibly simulated context if needed) for the signal, and then sampling observations.  Reasonable entity priors should produce a reasonable distribution of observations. 

When entities can be logically grouped into types, we can also drastically reduce the importance of $\pi$ and $\bd{\Pi}$ by warm-starting a new entity's reference vector distribution based on similar entities (e.g. other users for a new user).  We can sample reference vectors from the current posterior of similar entities, and use the empirical mean and covariance of those samples as the reference vector prior for the new entity.  Hence the hyperparameter priors would just be used for the initial entities of each type, and afterwards the observed data becomes influential.  We leave developing this idea further to future work.  

Specifying the dynamics is more difficult and likely problem-specific.  As rough guidance, we suggest that hyperparameters can again be shared across entities of the same type.  Then the memory can be intuitively set via considering the half-life of the dynamics.  Finally, let $\bd{\Omega}$, specified last, be a constant times the identity matrix.  These constants can be roughly determined via sampling reference and current vectors from the steady-state distribution, sampling observations for the reference signal and steady-state signal, and measuring the typical change in observation due to drift.

To summarize, this guidance involves setting these hyperparameters by considering answers to the following questions.  What is a typical reference observation?  What is a typical reference deviation in observation?  How long until an entity's parameters drift halfway back to their reference parameters in expectation?  What is a typical deviation in observation due to drift?  This guidance is a starting point, and analyzing a subset of data, perhaps repeatedly through cross-validation, could produce a better initialization.  Developing online solutions for fitting the hyperparameters is another area of future work. Future research could also consider online Kalman-filter-like algorithms for models that have latent variables, such as mixture and topic models.

\acks{CGU wants to thank Vijay Bharadwaj for suggesting the idea of extending the EKF to matrix factorization models, and Danny Ferrante and Nico Stier for supporting this work.  We also want to thank Ben van Roy and Yann Ollivier for useful feedback.}

\appendix
\section*{Appendix A: The Iterated Decoupled EKF}
\label{sec:alternativeUpdate}
Even if the DEKF is an adequate approximation to the EKF for factorization models, sometimes the EKF's approximate posterior is itself insufficiently accurate. Indeed, different second order approximations of $l(y)$ will result in update equations different from the EKF. Higher-order terms in the Taylor expansion about the prior mean may be relevant, especially if values of $\theta$ that are very likely according to the posterior are relatively far from $\mu.$ This suggests improving the accuracy of the EKF approximation by Taylor expanding about the MAP value of $\theta$, i.e., about the most likely value of $\theta$ according to the posterior. The iterated EKF (IEKF), described next, pursues this strategy. 

Consider approximating $l(y)$ about an arbitrary value $\bd{\gamma}$, rather than about $\mu$: 
\begin{align}
l(y) \approx & l(y, \bd{\gamma}) + \frac{\partial l(y)}{\partial \theta} |_{\bd{\gamma}} \big(\theta - \bd{\gamma} \big) + \frac{1}{2}
\big(\theta - \bd{\gamma} \big)'   \frac{\partial^2 l(y)}{\partial \theta^2}  |_{\bd{\gamma}}  \big(\theta - \bd{\gamma} \big). 
\nonumber
\end{align}
Working through the rest of the EKF derivation in the same way as before results in the following update equations:
 \begin{align} 
\bd{\Sigma}^{-1}_{\text{new}} = &  \bd{\Sigma}^{-1} - \frac{\partial^2 l(y)}{\partial \theta^2}  |_{\bd{\gamma}}, \label{eq:varUpdateDer2} \\
\bd{\delta} = & \bd{\Sigma}_{\text{new}}\biggr( \frac{\partial l(y)'}{\partial \theta} |_{\bd{\gamma}} 
- \frac{\partial^2 l(y)}{\partial \theta^2}  |_{\bd{\gamma}} \big(\bd{\gamma} - \mu \big)\biggr). \nonumber
\end{align}
Note that the column vector that multiplies $\bd{\Sigma}_{\text{new}}$ on the right to determine the mean update $\bd{\delta}$ now has two terms, and the second term goes to zero when $\bd{\gamma}=\mu.$  Also note that Equation~\ref{eq:varUpdateDer2} may lead to a ``covariance" that is not positive-definite.  Using the Fisher information matrix, like the EKF does, instead of the Hessian, is one alternative, and results in the update
 \begin{align} 
\bd{\Sigma}^{-1}_{\text{new}} = &  \bd{\Sigma}^{-1} + \bd{F}(\bd{\gamma}), \label{eq:varUpdateDer3} \\
\bd{\delta} = & \bd{\Sigma}_{\text{new}}\biggr( \frac{\partial l(y)'}{\partial \theta} |_{\bd{\gamma}} + \bd{F}(\bd{\gamma}) \big(\bd{\gamma} - \mu \big)\biggr). \label{eq:meanUpdateDer3}
\end{align}

Now consider the reference point $\bd{\gamma}$ that is self-consistent, i.e., that results in $\bd{\delta} = \bd{\gamma} - \mu$.  Under these circumstances, we get from Equation \ref{eq:meanUpdateDer3} that
\begin{align}
\frac{\partial l(y)'}{\partial \theta} |_{\bd{\gamma}} - \bd{\Sigma}^{-1}(\bd{\gamma} - \mu) = 0, \nonumber
\end{align}
where the left side is identical to the gradient of the log posterior evaluated at $\bd{\gamma}.$  Therefore, a self-consistent $\bd{\gamma}$ is a stationary point of the log posterior.  In particular, the MAP estimate of $\theta$ satisfies this equation.  The IEKF computes a MAP estimate by iterating 
\begin{align}
\bd{\gamma}_{\text{new}} = \bd{\gamma} + s (\bd{\Sigma}^{-1} + \bd{F}(\bd{\gamma}))^{-1}\left(\frac{\partial l(y)'}{\partial \theta} |_{\bd{\gamma}} - \bd{\Sigma}^{-1}(\bd{\gamma} - \mu)\right), \nonumber
\end{align} 
initialized from $\bd{\gamma} = \mu,$ using a line-search with step size $s \in [0, 1]$ to ensure that the log posterior is increasing on each iteration \citep[see][]{Skoglund7266781}. Upon convergence, the updated mean is $\bd{\gamma}$ and the updated covariance comes from Equation~\ref{eq:varUpdateDer3} evaluated at the converged $\bd{\gamma}$.\footnote{Typically it is the second-to-last $\bd{\gamma}$ that is used for the covariance, and the relevant terms have already been computed.}

After applying the Woodbury identity and some re-arrangement, $\bd{\gamma}_{\text{new}} - \bd{\gamma}$ can be written for our exponential family models as
\begin{align}
s \left[\mu - \bd{\gamma} + \bd{\Sigma} \frac{\partial \eta'}{\partial \theta} |_{\bd{\bd{\gamma}}}
\biggr[\bd{I} + \bd{B}(\bd{\gamma}) \biggr]^{-1} 
\bd{\Phi}^{-1}\biggr( y - h(\bd{\gamma}) + \bd{\Sigma}_{y} \bd{\Phi}^{-1} \frac{\partial \eta}{\partial \theta}|_{\bd{\bd{\gamma}}} (\bd{\gamma} - \mu) \biggr)\right]. \nonumber
\end{align}
This equation  can be evaluated similarly to Equation~\ref{eq:meanUpdate}.  The block-diagonal entity approximation to the covariance still implies that only parameters associated with entities in an observation are updated.  So our computational machinery can also be directly adapted for an iterated decoupled EKF.

\bibliography{main}

\end{document}